\documentclass[a4paper,fleqn]{cas-dc}
\usepackage[authoryear]{natbib}
\usepackage[T1]{fontenc}
\usepackage{booktabs}
\usepackage{array}
\usepackage{amsmath}
\usepackage{xcolor}
\usepackage{multicol}
\usepackage{listings}
\lstdefinestyle{md}{basicstyle=\def\baselinestretch{1}\ttfamily\footnotesize,columns=fullflexible,keepspaces=true,upquote=true,breaklines=true,postbreak=\mbox{\textcolor{gray}{$\hookrightarrow$}\space},frame=single,rulecolor=\color{gray},framesep=3pt,xleftmargin=3pt,xrightmargin=3pt,aboveskip=4pt,belowskip=4pt,escapeinside={(*}{*)}}
\newenvironment{rqlist}{\begin{list}{}{\small\setlength{\leftmargin}{2.8em}\setlength{\labelwidth}{2.4em}\setlength{\labelsep}{0.4em}\setlength{\itemsep}{1pt}\setlength{\parsep}{0pt}\setlength{\topsep}{3pt}}}{\end{list}}

\makeatletter
\def\@dbflt#1{\@ifnextchar[{\@xdblfloat{#1}}{\edef\reserved@b{\noexpand\@xdblfloat{#1}[\csname fps@#1\endcsname]}\reserved@b}}
\makeatother
\ExplSyntaxOn
\cs_set:Npn \__first_footerline:
  { \group_begin: \small \sffamily \__short_authors: :~ { \rmfamily \itshape Preprint } \group_end: }
\ExplSyntaxOff

\begin{document}

\shorttitle{Weight-space programming for tractor guidance controller}
\shortauthors{T. Oksanen}
\title [mode = title]{Weight-space programming for tractor guidance controller: large language models natively write neural network weights that steer a real tractor}
\author[1,2,3]{Timo Oksanen}\cormark[1]
\affiliation[1]{organization={Chair of Agrimechatronics, Technical University of Munich (TUM)}, country={Germany}}
\affiliation[2]{organization={Munich Institute of Robotics and Machine Intelligence (MIRMI), Technical University of Munich (TUM)}, country={Germany}}
\affiliation[3]{organization={Aalto University, Department of Electrical Engineering and Automation}, country={Finland}}
\cortext[1]{Corresponding author: timo.oksanen@tum.de}

\begin{abstract}
Weight-space programming is a process where a large language model reads and writes the weights of a neural network directly, guided by a natural language specification, without an optimiser or gradient descent. This study applies it to automatic tractor guidance, where a 253-parameter neural network steers along paths of lines and arcs and must be adapted when the machine changes, such as a slower steering actuator. GPT-6 Astra and Claude Fable 5.1 (also 5.2) edited the network for our research tractor using a small budget of simulated evaluations to repair a network trained for another tractor and write a controller into a network with zero weights. We compared them with an evolution strategy, gradient descent and pure pursuit in simulation and on the real tractor. Both models read the network and analysed the closed loop with control theory correctly, and Fable's repair without simulation succeeded only occasionally. With simulated feedback, both models repaired the trained network in single sessions and wrote a working controller from zero weights, whereas from zero the evolution strategy found none. The controllers written from zero were sparse and in a classical cascaded structure; one model edited the trained network while the other replaced it. Gradient descent reached higher accuracy with far more simulation, and all language model controllers steered the real tractor without intervention, with errors of the same order as pure pursuit. Language models can program the weights of a neural controller directly, much as control engineers design one. We call this Weight-Space Engineering.
\end{abstract}

\begin{keywords}
Weight-Space Engineering \sep WSE \sep Path tracking \sep Closed-loop control \sep Gradient-free optimisation \sep Control engineering
\end{keywords}

\maketitle

\section{Introduction}
\label{sec:introduction}

\subsection{Weight-space programming}
\label{sec:intro-wsp}

\begin{figure}[pos=b]
\centering\setlength{\tabcolsep}{3.5pt}
\includegraphics[width=\columnwidth]{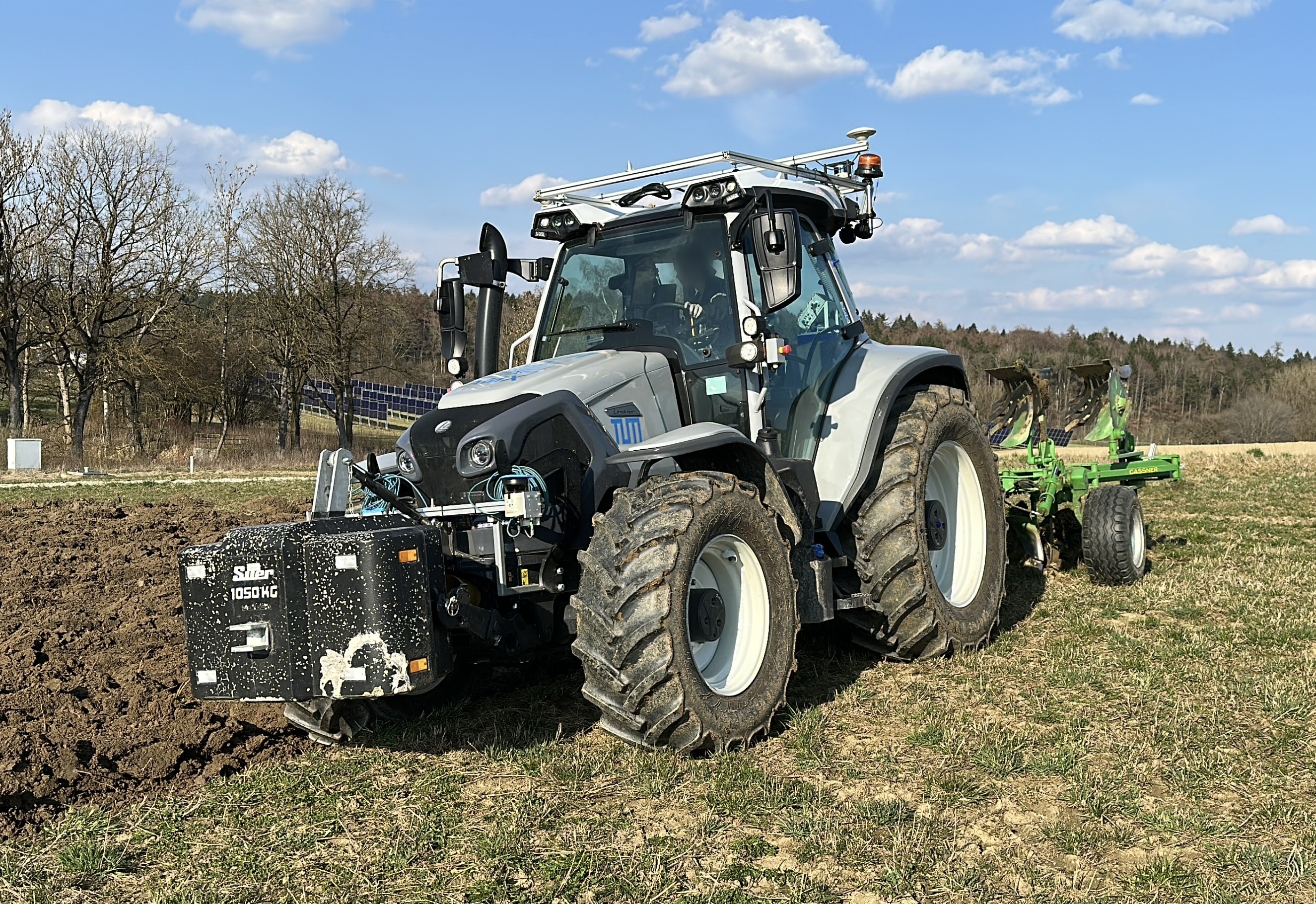}
\caption{The AMX G-trac tractor, which is the target machine of this study and whose steering dynamics define the simulated plant.}
\label{fig:tractor}
\end{figure}

\begin{figure*}[pos=tbp]
\centering
\includegraphics{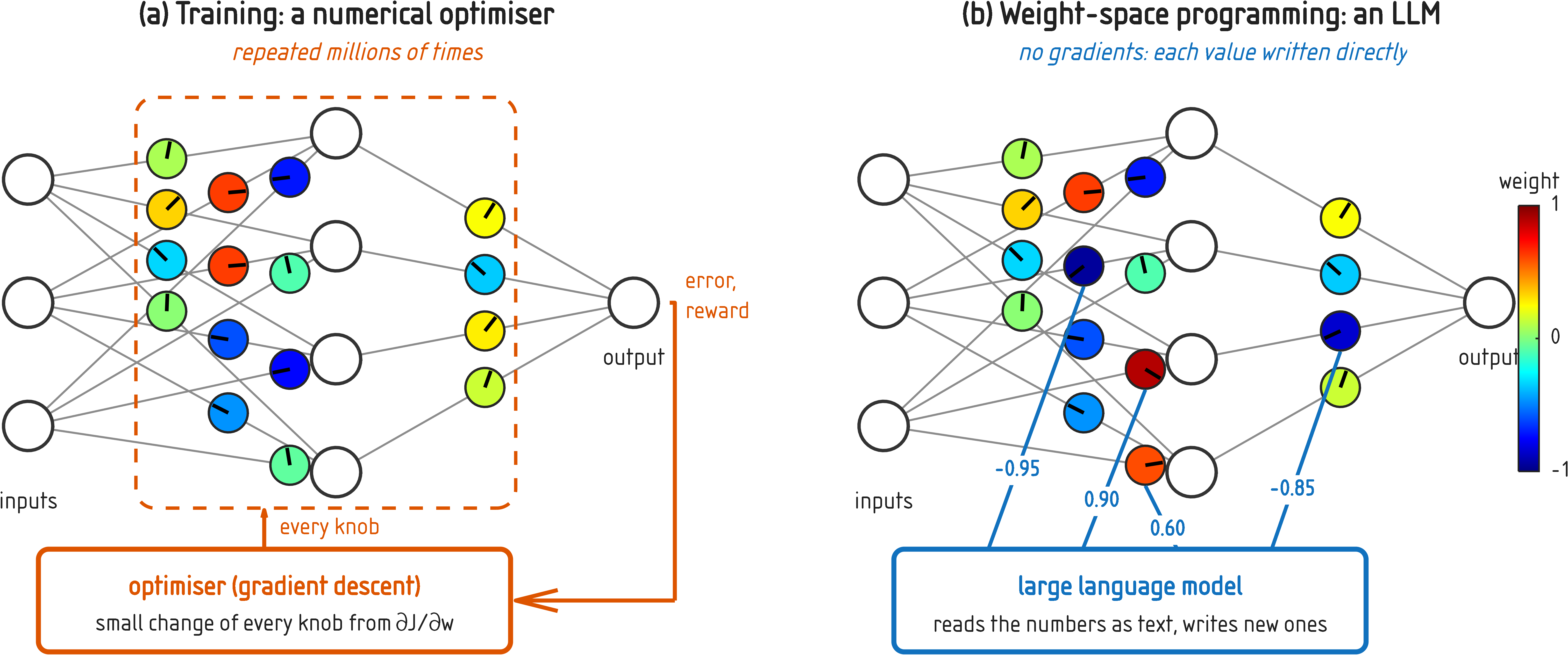}
\caption{Every weight of a neural network shown as a knob, where (a) training involves a numerical optimiser turning every knob a little from gradients and repeating this millions of times, while (b) weight-space programming involves an LLM reading the knob values as text and writing new values directly without gradients.}
\label{fig:knobs}
\end{figure*}

Weight-space programming is a process where a general-purpose large language model (LLM) reads and writes the weights of a neural network directly, guided by a natural-language specification instead of an optimiser or gradient descent~\citep{Takehi.2026}. The MIT Computer Science and Artificial Intelligence Laboratory (CSAIL) introduced the term weight-space programming in a research blog post on 2 November 2026~\citep{Takehi.2026}. This study employed the AMX G-trac tractor as the target machine (Fig.~\ref{fig:tractor}). To illustrate this concept, every weight of a network may be viewed as a knob (Fig.~\ref{fig:knobs}).

A human engineer chooses the network architecture, the data or the reward, and the loss, and a numerical optimiser then tunes every weight by gradient descent. In reinforcement learning, the gradient comes from the reward, for example with proximal policy optimisation (PPO)~\citep{Schulman.2017,Andrychowicz.2020}. The resulting controller works but remains a black box, where every weight is used and no one can explain why a single number has its value. A trained policy is tied to its plant, as a model-free policy learned for one actuator delay does not transfer to another delay~\citep{Chen.2020}. Therefore, adapting a trained network to a new machine means retraining it.

GPT-6 Astra repaired a failing reinforcement-learning policy for a car in the Kinetix physics environment in one reply, which changed 3.5~\% of the weights~\citep{Takehi.2026}. The model also wrote game-playing networks into all-zero weights, and these played better than networks trained with proximal policy optimisation for 100 million steps~\citep{Takehi.2026}. First follow-up studies verified discrete decision policies by model checking~\citep{Gross.2026} and debugged a robot policy in simulation~\citep{Zhang.2026}. However, all these results concern games, discrete tasks or simulation. This study addresses whether the same capability can serve the engineering of a controller for a physical machine, with feedback dynamics, limits and safety.

\subsection{Tractor guidance}
\label{sec:intro-guidance}

\begin{table*}[pos=b]
\centering
\caption{Eras of tractor guidance controllers by who writes the controller and in what form.}
\label{tab:eras}
\def\baselinestretch{1}\small
\begin{tabular}{@{}l>{\raggedright\arraybackslash}p{0.22\textwidth}>{\raggedright\arraybackslash}p{0.32\textwidth}>{\raggedright\arraybackslash}p{0.32\textwidth}@{}}
\toprule
Era & Who writes the controller & Form & Examples \\
\midrule
1 & Human & Control law, few hand-tuned parameters & Pure pursuit~\citep{Coulter.1992}, PID, LQR~\citep{OConnor.1997} \\
\addlinespace
2 & Human and numerical optimiser & Model and cost, action computed online & NMPC~\citep{Backman.2012,Soitinaho.2023} \\
\addlinespace
3 & Numerical optimiser & Network weights from data or reward & Deep reinforcement learning~\citep{Baby.2024,Pindl.2024} \\
\addlinespace
4 & Large language model & Network weights from understanding, no gradients & This paper \\
\bottomrule
\end{tabular}
\end{table*}

In this paper, automatic guidance means steering a tractor along a path with satellite positioning, such as GNSS or earlier GPS. This paper sorts tractor guidance controllers into eras by who writes the controller and in what form (Table~\ref{tab:eras}), where the eras overlap and do not replace each other. In Era 1, a human writes a control law with a few hand-tuned parameters. A tractor was steered automatically with centimetre-level GPS in 1997 by a state-feedback law whose gains were computed once offline from weights chosen by the engineer~\citep{OConnor.1997}. Autosteering systems typically use PID control on the cross-track error with gains set by trial and error~\citep{HeraudLange.2009}, and pure pursuit has the look-ahead distance as its single parameter~\citep{Coulter.1992,Snider.2009}. In Era 2, a human writes a model and a cost, and a numerical optimiser computes the action online from them, which includes nonlinear model predictive control of tractors and tractor-implement combinations~\citep{Backman.2012,Kayacan.2018,Soitinaho.2023,Moll.2026}, although the weights of its cost function still need manual tuning~\citep{Backman.2011}. Era 3 involves a numerical optimiser writing the weights of a neural network from data or reward, such as deep reinforcement learning on a simulated field robot~\citep{Baby.2024} and headland turns of the AMX G-trac learned with proximal policy optimisation and tested on the real tractor~\citep{Pindl.2024}. Mixing eras, for example nonlinear model predictive control with a neural network, is not a new era. Era 4 consists of an LLM writing the weights of a neural network from understanding, which this paper addresses.

\subsection{Weight-Space Engineering}
\label{sec:intro-wse}

Weight-space programming is the act of reading, explaining, repairing or writing weights directly for tasks such as games, text and images~\citep{Takehi.2026}. Programming is the act of writing code, whereas software engineering is the systematic, disciplined and quantifiable approach to developing, operating and maintaining software. This paper defines the term Weight-Space Engineering (WSE), which relates to weight-space programming as software engineering relates to programming and contains weight-space programming as one step. Weight-Space Engineering applies where the network acts in a technical system with physical consequences, for example a controller in a feedback loop with a plant. For a feedback controller, Weight-Space Engineering is control engineering, and this approach generalises to engineering in general. In WSE the human engineer works one level up and provides a specification of the plant, objective and limits, and gives knowledge of the field, such as control theory and intentions, but never the direction of an edit. The engineer does not explain what each number of the network means and may not know it, while the LLM grasps the network through the given theory and designs inside the network itself. Weight-Space Engineering keeps the parts of an engineering process and their costs, including evaluations, tokens, time and the engineer's effort. This process includes verification and validation through unseen tests, analysis with the theory of the field and a test on the machine, as well as deployment and maintenance, such as adapting a network to a changed machine. Traceability is maintained where the stated reasoning is checked against the result. In classical control engineering, the engineer designs the control law, whereas in WSE the engineer specifies, gives the theory, sets the evaluation budget and verifies, while the LLM designs. Table~\ref{tab:wse} lists each element of Weight-Space Engineering and how this study followed it.

\subsection{Objectives and contributions}
\label{sec:intro-objectives}

This study is the first worked instance of Weight-Space Engineering, where the prompt provides the plant and its limits and control theory as field knowledge. The study used an evaluation budget of 25 simulated evaluations and the exact linearised closed loop as the check for the retrofit of a trained network with 253 weights to a slower steering actuator. The overall question is whether a frontier LLM, without gradients, code execution or training, can engineer the weights of a small steering network for a tractor. The zero-shot test of RQ1 is weight-space programming without field knowledge; its prompt had no control-theory framing and no feedback evaluations. Both prompts provide the plant equations and the meaning of the network inputs and outputs, while only the prompts of RQ4 and RQ5 add the control-theory framing, whose effect the study did not isolate. The repair (RQ4) and the design from zero (RQ5) were single-session case studies with one session per model and start network, and the two repair sessions also served to develop the prompt.

\begin{rqlist}
\item[RQ1] Can a frontier LLM perform zero-shot repair of the trained network without feedback and without the control-theory framing?
\item[RQ2] Can the language model read the network by stating what its units compute and how its weights act in the closed loop?
\item[RQ3] Can the language model reason about the closed loop by using the principles of control theory?
\item[RQ4] Can the language model repair the trained network using a small evaluation budget of 25 simulated evaluations?
\item[RQ5] Can the language model write a working controller into a network whose weights are all zero?
\item[RQ6] Do the controllers written by the LLMs steer the real tractor in the field (Section~\ref{sec:field})?
\end{rqlist}

This study does not aim to develop the best possible guidance controller; nonlinear model predictive control already tracks paths of tractors and tractor-implement combinations~\citep{Backman.2012,Kayacan.2018,Soitinaho.2023,Moll.2026}, and beating such controllers is out of scope. This study does not benchmark the controllers written by the LLMs against all earlier guidance controllers; the classic pure pursuit controller serves as the baseline from the practice of guidance~\citep{Coulter.1992}. Other comparisons concern how the weights are written, namely search methods with the same evaluation budget and gradient descent on the same network. The central question is how an LLM can make a controller, not which controller is the most accurate.

The contributions of this paper are as follows: 1) this paper introduces Weight-Space Engineering and its first application to agricultural engineering and to a vehicle steering loop with actuator dynamics; 2) two frontier LLMs repaired a trained steering network for a slower actuator and wrote working controllers into an all-zero network, which this study compared with search methods that had the same evaluation budget; 3) this study checked the reading, the reasoning and the edits of the LLMs against the exact linearised closed loop, where the control laws that the LLMs stated matched the measured gains; 4) the designs of the LLMs are sparse, cascaded and readable and provide a good starting point for gradient descent; and 5) the controllers written by the LLMs also steered the real tractor.

\section{Related work}
\label{sec:related}

\subsection{Learned steering of vehicles}
\label{sec:rw-steering}

\citet{Folkers.2019} trained a car policy with PPO where the action included the steering angular velocity to prevent arbitrarily fast changes of the steering angle, using dense layers of 200 units and a convolutional branch for a perception map, and the policy also drove a real car. \citet{Weyrer.2024} learned path following and stabilisation of a multibody bicycle model with soft actor-critic, using an observation with the lateral offsets of points along the path ahead. \citet{Merton.2024} steered a simulated Formula Student car along cones with soft actor-critic and a steering-angle action using an actor with two hidden layers of 256 units, where a rate limit on the steering angle slowed learning, which did not converge at 90 degrees per second. \citet{Meyer.2020} followed paths and avoided collisions with an autonomous surface vessel using PPO with an actor of two hidden layers of 64 units. \citet{Caponio.2024} chose the derivative of the steering angle as the action to obtain smooth steering and tracked paths with a scaled robotic car in experiments. \citet{Jose.2023} commanded the rudder angle of a container ship with a rate-limited rudder, and the policy needed a higher control effort than a PD controller.

\citet{Nath.2021} showed that a state augmented with the actions sent during the delay turns a delayed decision process into an undelayed one. \citet{Ramstedt.2020} showed that a buffer of the last sent actions, covering the longest delay, is needed for the Markov property under random delays. \citet{Peng.2018} transferred a robot policy from simulation to a real robot with randomised dynamics, where the success rate fell from 0.89 to 0.29 without randomised latency.

Linear and small policies are competitive in continuous control~\citep{Mania.2018,Rajeswaran.2017}. Equivariant policies learn better than policies trained with data augmentation~\citep{Wang.2022,vanderPol.2020}, and robot policies employ mirror symmetry~\citep{Mittal.2024}. However, gradient descent trained all these networks, and no study has edited the weights of a trained steering network directly.

\subsection{Weight-space programming}
\label{sec:rw-wsp}

\citet{Takehi.2026} used one LLM, GPT-6 Astra, in eight experiments that read, repaired or wrote networks, which ranged from about 17 000 weights up to language models. Some experiments gave the LLM no tools, and others provided a Python workspace and a budget of forward passes of the network. The repaired car policy of a physics game had 171 664 weights, and its repair took one reply without tools. The authors call the results very early and state that editing removed an obstacle that hid an ability but could not add a missing ability. Most experiments were single runs, and the comparisons were PPO references and networks trained from data, rather than non-LLM search with the same evaluation budget. \citet{Lindner.2023} previously compiled programs written in a formal language into transformer weights for interpretability research, whereas weight-space programming starts from a natural language specification that a general LLM reasons about.

After the post on 2 November 2026, the follow-up studies appeared as arXiv preprints between 3 and 12 November 2026. \citet{Gross.2026} employed two LLMs to write policies for discrete Markov decision processes, which a probabilistic model checker verified and which were optimal in all 18 runs after the first writing step. \citet{Zhang.2026} used GPT-6 Astra with a Python workspace and a budget of simulator rollouts to edit 65 embedding rows of a robot policy with 3.6 billion parameters, where success on paraphrased instructions rose from 86.3 to 92.0~\%, and the authors report no transfer to physical robots. \citet{Nusrat.2026} used Claude Opus 5.5 without tools to audit and repair clinical decision networks of 433 parameters with a median of 3 changed parameters, and the model also wrote networks from all-zero weights. \citet{Sato.2026} used Claude Opus 5.5 without tools to write networks of at most 501 parameters that solve partial differential equations with 5 rounds of feedback, which were exact only where the model built a known analytical solution into the network. \citet{Liu.2026} found that GPT-6 Astra without tools read networks of 353 weights reliably, whereas its reading degraded at 3457 weights. \citet{Lehmann.2026} observed that GPT-6 Astra rebuilt the policy in the repair of trained game policies, whereas Claude Opus 5.5 made small edits. \citet{Salloum.2026} developed a non-LLM optimisation method designed from the published logs of the agent of \citet{Takehi.2026}, which exceeded the repair of a language model by that agent with fewer forward passes. None of these studies concerned a continuous plant with actuator dynamics, a closed-loop analysis or a physical machine.

Earlier, \citet{Kevian.2024} tested GPT-4, Claude 3 Opus and Gemini 1.0 Ultra on 147 undergraduate control problems, where Claude 3 Opus solved 58.5~\% directly and 68.7~\% after checking its own answers, and all three models misread Bode and Nyquist plots. \citet{Narimani.2025} let LLM agents choose the gains of P, PI, PD, PID or full-state-feedback laws from performance metrics of simulations or hardware trials, with at most 4 gains per law. \citet{Yang.2024} used LLMs as optimisers that proposed new solutions from a history of scored solutions, and their numerical tests involved linear regression with two variables and travelling-salesman problems of up to 50 nodes, where performance degraded sharply with the problem size. These studies had LLMs solve control problems on paper, tune a few gains of a given control law~\citep{Guo.2024} or optimise a few variables, whereas in weight-space programming the LLM writes the internal weights of a neural network. \citet{Bosio.2024} let an LLM write control policies as Python code rather than network weights, and an evolutionary search improved the programs over about 10 000 candidates per run for simulated pendulum swing-up and ball-in-cup tasks.

\section{Materials and methods}
\label{sec:methods}

\subsection{Tractor and steering actuator model}
\label{sec:plant}

This study used the AMX G-trac research tractor (Fig.~\ref{fig:tractor}), which has four-wheel steering and whose rear axle was held straight. The simulation model is a kinematic bicycle with the reference point on the rear axle, a constant speed of 1 m/s, and a time step of 0.05 s. The steering chain consists of four stages, where the controller integrates a steering-rate command of at most 15 deg/s into a command angle clipped at 30 deg. A pure dead time of 0.45 s and a first-order lag with a time constant of 0.75 s follow, and the physical wheel limits are 34 deg and 26.5 deg/s. The dead time and lag are rounded to the time step of 0.05 s from the identification values of 0.43 s and 0.73 s. The wheelbase is 2.42 m, and Table~\ref{tab:plant} lists all values of the plant, the paths, the network and the training recipe.~\citep{Soitinaho.2026,Hefele.2024}

\begin{table}[pos=htbp]
\centering
\def\baselinestretch{1}\footnotesize
\caption{Simulated plant, paths, network and the PPO recipe of the start network S1. The plant values are those of the AMX G-trac simulation used in all sessions and evaluations.}
\label{tab:plant}
\begin{tabular}{@{}>{\raggedright\arraybackslash}p{0.45\columnwidth}>{\raggedright\arraybackslash}p{0.50\columnwidth}@{}}
\toprule
Parameter & Value \\
\midrule
\multicolumn{2}{@{}l}{\textit{Plant: AMX G-trac simulation}} \\
Model & kinematic bicycle, reference point on the rear axle \\
Wheelbase $L$ & 2.42 m \\
Speed $v$ & 1 m/s \\
Time step & 0.05 s \\
Steering-rate command, $u = \pm 1$ & $\pm$15$^\circ$/s \\
Command angle clip (controller) & $\pm$30$^\circ$ \\
Dead time $T_\mathrm{d}$ & 0.45 s (9 steps) \\
Lag time constant $\tau$ & 0.75 s \\
Wheel angle and rate limits & $\pm$34$^\circ$, $\pm$26.5$^\circ$/s \\
\addlinespace
\multicolumn{2}{@{}l}{\textit{Paths}} \\
Length, segments & 60 m, 10 lines and arcs \\
Curvatures & 0, $\pm$0.25, $\pm$0.5, $\pm$0.75, $\pm$1 times 0.1848 1/m (minimum radius 5.4 m) \\
Preview & 4 chunks of at most 4 m \\
Failure & lateral error above 2 m \\
\addlinespace
\multicolumn{2}{@{}l}{\textit{Network}} \\
Token encoder (shared) & 9$\to$4$\to$4, tanh \\
Head & 22$\to$8$\to$1, tanh hidden layer \\
Head inputs & 4$\times$4 token features, wheel angle, command angle, last 4 outputs \\
Output & mirror-symmetric, $u = \tanh(\tfrac{1}{2}(f(o) - f(\bar{o})))$, $\bar{o}$: mirrored observation \\
Parameters & 253 \\
\addlinespace
\multicolumn{2}{@{}l}{\textit{PPO training of the start network S1}} \\
Training plant & $T_\mathrm{d}$ = 0.2 s, $\tau$ = 0.2 s, $L$ = 2 m \\
Environment steps & 20 million \\
Parallel environments, rollout length & 128, 128 steps \\
Epochs, minibatch & 5, 4096 \\
Learning rate & 0.001, linear decay to 0 \\
Discount, GAE parameter & 0.995, 0.95 \\
Clip ratio, value coefficient & 0.2, 0.5 \\
Entropy coefficient, gradient norm limit & 0, 0.5 \\
Forced curvature reversals & probability 0.08 per junction \\
Step reward & $r = 0.1\,(r_y r_\psi - 0.02\,u^2)$, 1 less at failure \\
Lateral term $r_y$ & $0.5\,e^{-(e_y/0.05)^2} + 0.5\,e^{-(e_y/0.3)^2}$, $e_y$ in m \\
Heading term $r_\psi$ & $0.8 + 0.2\,e^{-(e_\psi/0.15)^2}$, $e_\psi$ in rad \\
Critic & two tanh layers of 256; the inputs of the network plus the true $e_y$ and $e_\psi$ \\
Seeds & 3, best on its own training plant \\
\bottomrule
\end{tabular}
\end{table}

\subsection{The 253-parameter steering network}
\label{sec:network}

\begin{figure}[pos=htbp]
\centering\setlength{\tabcolsep}{3.5pt}
\includegraphics[width=\columnwidth]{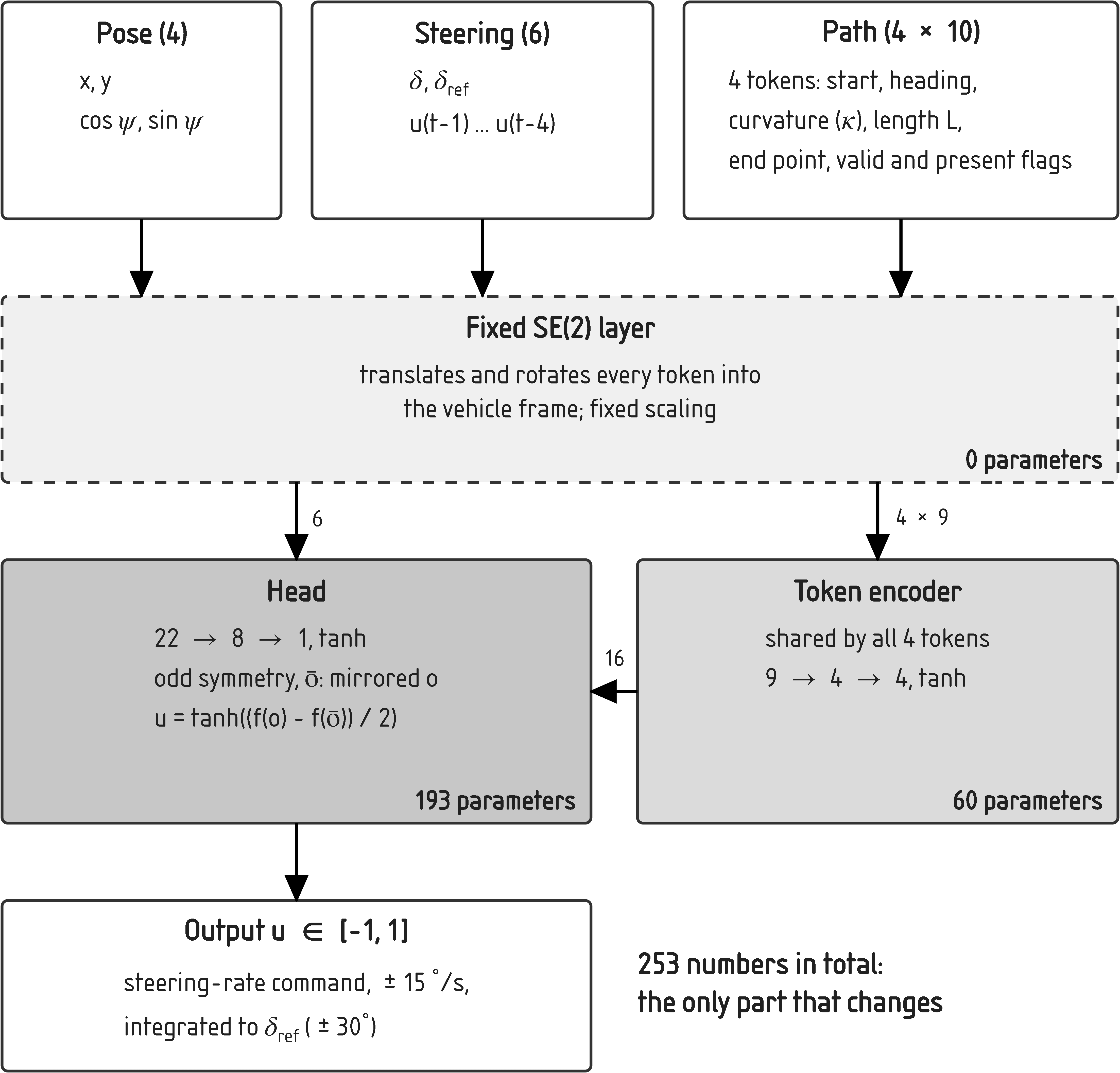}
\caption{The steering network with 253 parameters and the fixed SE(2) layer, the shared token encoder with 60 parameters and the head with 193 parameters, where the architecture is fixed and only the 253 numbers change. The pose and the path chunks enter only the SE(2) layer, which passes the 4 tokens of 9 features in the vehicle frame to the token encoder.}
\label{fig:network}
\end{figure}

Every 0.05 s the network receives the pose of the tractor, the steering state and the path ahead, where the steering state comprises the measured wheel angle, the command angle and the last 4 rate commands. The path is cut into chunks of at most 4 m, and the network sees the current chunk and the next 3. Each segment of the path is split into the smallest number of equal chunks that are not longer than 4 m. The current chunk is the chunk on which the projection of the rear axle lies, so that the first token keeps its start point until the rear axle passes the end of its chunk. A fixed SE(2) layer without parameters turns each chunk into a token of 9 features in the vehicle frame. A token encoder 9-4-4 with tanh units and 60 parameters is shared by the 4 tokens. A head 22-8-1 with a tanh hidden layer and 193 parameters gives the output, which is a steering-rate command in the range from -1 to 1 scaled to 15 deg/s. The output has an exact mirror symmetry, $u = \tanh(0.5\,(f(o) - f(\mathrm{mirror}(o))))$, which needs no extra parameters; therefore, the policy is odd and the output bias has no effect. Fig.~\ref{fig:network} shows the blocks of the 253 parameters.

A preliminary stage of this work trained end-to-end steering networks by pure reinforcement learning via proximal policy optimisation on a nominal plant with dead time 0.2 s, lag 0.2 s and wheelbase 2.5 m. The tokenised path was inspired by vectorised map encoders~\citep{Gao.2020} and set networks~\citep{Zaheer.2017}. This study selected the smallest network that completed all 512 held-out paths without failure, which had 253 parameters.

The architecture was proven in the preliminary stage and fixed before any LLM session. Gradient descent with the same structure reached a good controller on the AMX G-trac (Section~\ref{sec:res-search}), which implies that a shortfall of the LLMs lies in the weights and not in the architecture. The frozen structure makes the study a pure weight-space test.

\subsection{Paths, objective and data splits}
\label{sec:objective}

Each path is 60 m long with 10 segments, starting at the origin and heading along $x$, where segment 0 is straight. The other segments are lines or arcs whose curvature comes from the symmetric set 0, plus or minus 0.25, 0.5, 0.75 and 1 times 0.1848 1/m, providing a minimum radius of 5.4 m. Segment lengths are random with a minimum of 2 m, so that the curvature jumps at the junctions; a full reversal from left to right is the hardest case. Each of the 10 segment lengths is 2 m plus a share of the remaining 40 m, where shares are drawn uniformly over all possible splits and every curvature level has the same probability. The three path sets contain no forced curvature reversals, and only the training with proximal policy optimisation used forced reversals. An episode fails when the lateral error exceeds 2 m.

The objective $J$ is the mean over paths of the RMS lateral error in cm plus 200 times the share of failed paths, where a lower value is better. This study used three fixed path sets generated with fixed seeds, including a feedback set of 64 paths, which constitutes the only data the LLMs saw. A select set of 256 paths serves only for choices among runs, such as the best run of proximal policy optimisation. Finally, a held-out set of 512 paths provides every reported score.

\subsection{Start points: the trained network S1 and the zero network S0}
\label{sec:starts}

The study used two start points for the LLM sessions, where S1 is a network trained for another tractor using proximal policy optimisation with the same recipe, a wheelbase of 2.0 m, and a faster actuator with dead time 0.2 s and lag 0.2 s. The command history of the network covers the last 4 steps, or 0.2 s, which matches the dead time of the nominal plant but covers less than half of the 0.45 s dead time of the AMX G-trac. This training used 3 seeds, and the study selected the best on the select set of its own plant. S1 is a new network with the architecture of the preliminary stage, whereas the earlier network of the preliminary stage had a wheelbase of 2.5 m and served only the reading and control-theory tests (Section~\ref{sec:understanding-methods}). On its own plant, S1 reached $J = 1.36$ on the select set, whereas on the AMX G-trac it reached $J = 199.5$, with 186 of 256 select paths failed. S0, the zero network, has all 253 values zero, which means its output is exactly zero and every gradient of proximal policy optimisation is zero; therefore, gradient descent cannot start from S0. S1 asks for a repair and S0 asks for a design from nothing.

\subsection{Large language models and API access}
\label{sec:llms}

This study used two frontier LLMs, GPT-6 Astra from OpenAI and Claude Fable 5.1 from Anthropic, which were accessed through their vendors' public APIs (Table~\ref{tab:api}). The models had no tools, including no code execution, no file access and no web search, and the harness ran every simulation. The requests contained no tools field, which the session scripts checked; furthermore, the study saved all requests and responses verbatim. The four sessions ran on 7 November 2026, and Table~\ref{tab:sessions} provides their turns, time and tokens.

\begin{table}[pos=htbp]
\centering\setlength{\tabcolsep}{3.5pt}
\def\baselinestretch{1}\small
\caption{Access to the two LLMs in the four sessions. Both received identical message texts and no system prompt.}
\label{tab:api}
\begin{tabular}{@{}>{\raggedright\arraybackslash}p{0.27\columnwidth}>{\raggedright\arraybackslash}p{0.3\columnwidth}>{\raggedright\arraybackslash}p{0.37\columnwidth}@{}}
\toprule
 & GPT-6 Astra & Claude Fable 5.1 \\
\midrule
Vendor & OpenAI & Anthropic \\
Model identifier & \texttt{gpt-6-astra} & \texttt{claude-fable-5-1} \\
Interface & Responses API & Messages API, streaming \\
Reasoning & effort high, summary auto & thinking always on, effort high \\
Output tokens per turn, maximum & 100 000 & 128 000; 64 000 in turn 1 of S1; lower in late turns of S0 (cost cap) \\
Conversation state & chained on the server & full conversation re-sent each turn, prompt caching \\
Tools & none & none \\
\bottomrule
\end{tabular}
\end{table}

\begin{table*}[pos=tbp]
\centering
\def\baselinestretch{1}\small
\caption{Statistics of the four LLM sessions, one per model and start point, each with an evaluation budget of 25. API time is the sum of the response times of the API calls. An invalid proposal used its turn and received no score.}
\label{tab:sessions}
\begin{tabular}{@{}lrrrr@{}}
\toprule
 & S1 Astra & S1 Fable & S0 Astra & S0 Fable \\
\midrule
Turns & 25 & 25 & 25 & 25 \\
Valid proposals & 24 & 24 & 22 & 23 \\
API time, total [s] & 934 & 2512 & 1676 & 4157 \\
API time per turn, median [s] & 25 & 62 & 53 & 56 \\
API time, first turn [s] & 168 & 868 & 252 & 1435 \\
Output tokens [thousand] & 70 & 197 & 78 & 322 \\
of which reasoning [thousand] & 30 & 154 & 47 & 279 \\
\bottomrule
\end{tabular}
\end{table*}

\subsection{LLM sessions: protocol, prompt and evaluation budget}
\label{sec:protocol}

This study used one prompt template where the text was byte-identical for both models and used plain ASCII (Appendix~\ref{app:prompt}). The prompt began with the working conditions, which specified a text-only exchange without code execution and required all analysis by the model's own reasoning. It then described the plant with its equations, the provenance of the start network, the forward pass in five steps, the rules and the session protocol. The data followed, including the labelled weight tables, the weights in JSON, and the metrics of the start network with one recorded episode. The control-theory framing allowed the model to reason with control engineering methods, such as linearisation about straight driving, feedback gains, closed-loop poles and damping, margins and describing functions. The prompt gave no hint about the direction of the edit. The rules required that the model change only the numbers while keeping keys and shapes, using finite values with an absolute value of at most 10.

This study defined the evaluation budget as the number of feedback evaluations that a session may use, which was at most 25. One feedback evaluation consisted of the simulation of one proposal on the 64 feedback paths. After each proposal, the model received feedback comprising the metrics, one recorded episode around the worst error, and a table of all proposals so far. An invalid proposal, such as one that was unparsable, had the wrong shape, or had a value above 10 in magnitude, used its turn and received no score. For the LLMs and the search baselines alike, the controller at an evaluation budget of $n$ was the candidate with the lowest feedback objective among the start network and the first $n$ proposals, which the study scored once on the held-out set to produce a performance curve. The study developed the prompt in pilot sessions on S1 and froze it afterwards. However, the API cost allowed one session per model and start point, so the pilot sessions on S1 served as the S1 case study\footnote{In the S1 session of GPT-6 Astra, proposal 2 was scored invalid because of a parser error, and GPT-6 Astra resubmitted it unchanged as proposal 3. In the S1 session of Claude Fable 5.1, turn 1 reached an output cap of 64 000 tokens, so this session has no proposal at an evaluation budget of 1.}, with one session per case and no statistical tests.

The APIs return the hidden reasoning of the models only as summaries written by the provider with an undisclosed method, and this study saved these summaries verbatim with the replies. Summaries exist for Fable 5.1 from turn 5 of its session from zero, when the study switched them on, and for GPT-6 Astra in 3 of the 25 turns of its session from zero.

\subsection{Zero-shot, reading and control-theory tests}
\label{sec:understanding-methods}

The zero-shot test of RQ1 used the task of the S1 sessions with no evaluations, which involved one proposal without simulation or feedback. The prompt consisted of the frozen session prompt without the control-theory framing. This study performed 10 repetitions, where each repetition used a new Claude Fable 5.1 agent that had only two tools, reading a file and writing a file, and no tool to list or search folders. Each agent listed the files it read, and the harness recorded the number of tool calls, which matched the 4 files of its own folder and the 1 proposal that it wrote. Success was defined before the runs as no failure on the held-out paths and a $J$ value below that of the start network.

The other tests used the earlier network trained on the nominal plant with a wheelbase of 2.5 m and a pilot variant of the plant, where both models received identical questions. The reading test asked for the units of the inputs, the linearised feedback gains and an explanation of the forward pass, which this study scored against the exact gains from automatic differentiation. The closed-loop test asked for stability, crossover frequencies, the phase margin, the stable range of a loop-gain factor, limit cycles by describing functions and a gain design on four plants, and it was scored against the exact discrete linear model. The per-parameter test asked for the direction of the effect on stability and on tracking for each row of the network, which was scored against measured sensitivities. Each model took each test once, where GPT-6 Astra operated through the API and Claude Fable 5.1 acted as an agent with file access restricted to its own folder. The agent read the prompt and the weights from files in its folder and wrote its answer to a file, whereas the 25-turn sessions of Fable 5.1 ran over the API without tools (Section~\ref{sec:llms}), and only the tests of this section used the agent.

\subsection{Baselines: gain grid, evolution strategy, pure pursuit and gradient descent}
\label{sec:baselines}

The gain grid sampled one output gain on a geometric grid from 0.1 to 10, visited from 1 outwards. A (1+1) evolution strategy operated on all 253 weights and employed a 1/5 success rule for the step size, using 5 seeds with 25 evaluations and a long version with 1000 evaluations. Each step of the evolution strategy added Gaussian noise to every weight, scaled by the step size times the weight magnitude plus 0.05, where the step size started at 0.1 and grew by a factor of 1.5 after an improvement or shrank by the fourth root of 1.5 otherwise; the weights had no bounds. On S0, the evolution strategy also started from the noisy zero networks that PPO received (Section~\ref{sec:gd-from-llm}), using noise of 0.01 and 0.05 across 5 seeds each with 1000 evaluations. This study scored both baselines with the same evaluation budget and selection rule as the LLMs.

This study employed three reference controllers, where PP used pure pursuit~\citep{Coulter.1992,Snider.2009} with a look-ahead of 4.0 m tuned on the feedback set of the AMX G-trac over 12 values from 1.5 to 7.0~m. The GD-T controller was trained with proximal policy optimisation from scratch on the AMX G-trac using the recipe of the preliminary stage, which involved 20 million steps and 6 seeds to select the best on the select set. GD-1 is the best reference trained by proximal policy optimisation, which was the best on the select set among all such runs on the AMX G-trac and turned out to be a fine-tuning of S1.

\subsection{Gradient descent from the LLM designs}
\label{sec:gd-from-llm}

This study started proximal policy optimisation with the GD-T recipe from each LLM controller, from S1, and from the zero network. The study added noise of 0, 0.01 or 0.05 to weights that were exactly zero, using 3 to 5 seeds per start; for the zero network, 98 seeds per noise level were used. Noise was needed for the zero network because a network that is exactly zero cannot learn. AS-0-GD represents the best run on the select set among the runs started from Astra's design from zero.

\subsection{Mechanism analysis}
\label{sec:mechanism}

This study linearised each network controller about straight driving by automatic differentiation to determine gains on the lateral error, heading error, wheel angle, command angle and command history. The operating point was a straight path cut into chunks of 4 m, where the rear axle stood 2 m into the current chunk and a lateral or heading error moved the start and end points of all 4 tokens in the vehicle frame. These gains were closed with the exact discrete linear plant, which included the lag, the command integrator, the kinematics and a dead time as a delay of 9 steps; saturations and clips were excluded so that the analysis remained valid for small deviations. The study reported the stability, the gains and the frequency and damping ratio of the dominant oscillatory mode. The control laws that the LLMs stated were compared with the measured gains, and this post-run analysis was never shown to the LLMs.

\section{Results}
\label{sec:results}

\subsection{Overview}
\label{sec:res-overview}

With simulated feedback, both LLMs repaired the trained network S1 and wrote a working controller into the all-zero network S0. All four final controllers of the LLMs had a stable linearised loop and completed the unseen test paths, where three completed them without a failure and the repair by Fable had 1 failure in 512 paths, whereas the search with the same evaluation budget worked only from S1 and never from S0. Without simulated feedback, the zero-shot repair worked in two of ten attempts; Table~\ref{tab:heldout} and Table~\ref{tab:loop} show all numbers.

\begin{table*}[pos=tbp]
\centering
\def\baselinestretch{1}\footnotesize
\caption{Held-out performance of the controllers on the AMX G-trac simulation plant (512 unseen random paths). $J$: mean RMS lateral error in cm plus 200 times the share of failed paths (lower is better); a path fails when the lateral error exceeds 2~m. Failures: out of 512 paths. Max.: largest lateral error in cm (a failed path ends at 2~m, so a controller with failures shows about 200). Non-zero: weights out of 253. Budget: simulated evaluations (LLM sessions and search) or PPO time steps (M: million). Groups of runs (number of runs in brackets): median (min--max) of $J$, range of failures with the number of runs without failures in brackets, median of the maximum error.}
\label{tab:heldout}
\setlength\tabcolsep{2.5pt}
\begin{tabular}{@{}lrrrrr@{}}
\toprule
Controller & Budget & $J$ & Failures & Max. & Non-zero \\
\midrule
\multicolumn{6}{@{}l}{\textit{Start points}} \\
S1 start & -- & 190.53 & 352 & 204 & 252 \\
S0 start (zero weights) & -- & 268.09 & 512 & 204 & 0 \\
\midrule
\multicolumn{6}{@{}l}{\textit{Repair of S1 (zero-shot: RQ1)}} \\
AS-1 (Astra) & 25 & 7.82 & 0 & 87 & 220 \\
FA-1 (Fable) & 25 & 9.36 & 1 & 201 & 25 \\
Fable zero-shot, run 6 & 0 & 12.65 & 0 & 149 & 252 \\
Fable zero-shot, run 9 & 0 & 49.06 & 0 & 161 & 252 \\
Fable zero-shot, 8 others & 0 & 119.95 (47.38--262.76) & 21--501 & 203 & \\
Gain grid & 25 & 97.76 & 170 & 204 &  \\
(1+1)-ES (5) & 25 & 15.41 (9.37--37.85) & 1--28 & 202 & \\
(1+1)-ES (5) & 1000 & 10.00 (5.07--10.24) & 1--9 & 202 & \\
\midrule
\multicolumn{6}{@{}l}{\textit{Synthesis from zero weights (S0)}} \\
AS-0 (Astra) & 25 & 4.94 & 0 & 60 & 53 \\
FA-0 (Fable) & 25 & 7.00 & 0 & 134 & 50 \\
Gain grid & 25 & 268.09 & 512 & 204 &  \\
(1+1)-ES (5) & 25 & 267.95 (264.30--268.11) & 512 & 204 & \\
(1+1)-ES (5) & 1000 & 257.88 (255.41--268.10) & 512 & 205 & \\
(1+1)-ES, noise 0.01 (5) & 1000 & 261.47 (255.41--268.14) & 512 & 205 & \\
(1+1)-ES, noise 0.05 (5) & 1000 & 268.10 (268.09--268.13) & 512 & 204 & \\
\midrule
\multicolumn{6}{@{}l}{\textit{References}} \\
Pure pursuit (look-ahead 4 m) & -- & 6.98 & 0 & 38 &  \\
GD-T (PPO from scratch) & 20\,M & 3.43 & 0 & 130 & 252 \\
GD-1 (PPO from S1) & 20\,M & 2.78 & 0 & 56 & 253 \\
AS-0-GD (PPO from AS-0) & 20\,M & 3.13 & 0 & 90 & 253 \\
PPO from zero, noise 0.01 (98) & 20\,M & 5.05 (3.82--7.91) & 0--6 (48) & 200 & \\
PPO from zero, noise 0.05 (98) & 20\,M & 4.91 (3.71--8.20) & 0--4 (38) & 200 & \\
\bottomrule
\end{tabular}
\end{table*}

\begin{table}[pos=htbp]
\centering\setlength{\tabcolsep}{3.5pt}
\def\baselinestretch{1}\footnotesize
\caption{Linearised closed loop of the controllers on the AMX G-trac simulation plant at straight driving. $f$: frequency of the dominant oscillatory mode; $\zeta$: its damping ratio. The loop of S1 start is marginally unstable ($\zeta = 0$); all other loops are stable. Gains of the network output before the $\tanh$: outer loop on the lateral error $e_y$ (per m) and the heading error $e_\psi$ (per rad), inner loop on the measured wheel angle $\delta$ and the command angle $\delta_\mathrm{ref}$ (per rad).}
\label{tab:loop}
\setlength\tabcolsep{1.6pt}
\begin{tabular}{@{}lrrrrrr@{}}
\toprule
Controller & $f$ [Hz] & $\zeta$ & $K_y$ & $K_\psi$ & $K_\delta$ & $K_{\delta_\mathrm{ref}}$ \\
\midrule
S1 start & 0.120 & $0.00$ & $-2.40$ & $-6.81$ & $-2.00$ & $-2.36$ \\
\addlinespace
AS-1 (Astra, repair) & 0.187 & $0.37$ & $-1.30$ & $-10.39$ & $-5.05$ & $-4.60$ \\
FA-1 (Fable, repair) & 0.072 & $0.62$ & $-1.43$ & $-5.85$ & $-2.86$ & $-2.86$ \\
AS-0 (Astra, from zero) & 0.141 & $0.44$ & $-1.73$ & $-15.74$ & $-3.82$ & $-7.64$ \\
FA-0 (Fable, from zero) & 0.119 & $0.37$ & $-1.64$ & $-16.79$ & $-1.60$ & $-9.85$ \\
\addlinespace
GD-T (PPO from scratch) & 0.059 & $0.63$ & $-2.11$ & $-7.96$ & $-3.50$ & $-3.76$ \\
GD-1 (PPO from S1) & 0.116 & $0.26$ & $-2.33$ & $-9.66$ & $-2.98$ & $-4.23$ \\
AS-0-GD (PPO from AS-0) & 0.077 & $0.44$ & $-1.83$ & $-5.92$ & $-2.17$ & $-3.38$ \\
\bottomrule
\end{tabular}
\end{table}

\subsection{Zero-shot repair without simulation}
\label{sec:res-zeroshot}

Ten independent Fable agents each provided one proposal without simulation or feedback and without the control-theory framing in the prompt (RQ1). Two of the ten proposals repaired the network, which resulted in no failures on the unseen paths and a lower error than the start (Table~\ref{tab:heldout}). One of these agents tracked the paths with about twice the error of pure pursuit, while the other completed the paths but tracked loosely. Seven of the ten proposals improved on the start, whereas three made it worse.

All ten replies named the same cause from the recorded episode, where the measured wheel angle lags the command angle, resulting in wind-up and a limit cycle. These replies used control engineering reasoning, such as linearisation, phase margin and a Smith predictor, although the prompt did not ask for it. The linearised loop of the start S1 was unstable on the AMX G-trac plant (Table~\ref{tab:loop}), and two proposals left it unstable. The damping ratio of the linearised loop of each proposal ordered the outcomes almost perfectly, showing a rank correlation of $-$0.98 with the number of failed paths. The two proposals that worked were among those that damped the linear loop most.

\subsection{Reading the network}
\label{sec:res-reading}

This study tested the earlier nominal network of the pilot with each model once (RQ2, Section~\ref{sec:understanding-methods}), where both models computed the linearised feedback gains by hand to two or three digits, which remained inside their own stated uncertainty intervals. Both models identified 89 of 96 signs of the input effects per hidden unit correctly, and Fable computed the hidden-unit activations within 0.001. Astra declined the per-parameter table for 252 of 253 rows, stating that the closed-loop effect of one weight cannot be judged locally.

\subsection{Reasoning with closed-loop control theory}
\label{sec:res-ctrl}

Both models performed closed-loop analysis on four plants, where they correctly identified every stability verdict, every stable gain range, and the existence of every limit cycle (RQ3). The median relative error of the crossover frequencies was about 1~\% for GPT-6 Astra, whereas it was about 6~\% for Claude Fable 5.1. Both models designed gains for a damping ratio of at least 0.5, and these designs met the requirement in the exact linear loop. A per-weight test asked for the direction of the effect of each of the 253 weight rows on stability; both models abstained on the same 5 rows, the curvature weights of the first encoder layer and the output bias, which have no effect in the loop linearised about straight driving. This left 248 scored rows, where GPT-6 Astra identified 248 of 248 correctly, whereas Claude Fable 5.1 identified 227 of 248. This study shows that both models analysed the closed loop correctly, although GPT-6 Astra did so more precisely.

\subsection{Repair with simulated feedback}
\label{sec:res-s1}

For RQ4, the start S1 failed on 352 of 512 unseen paths (Table~\ref{tab:heldout}), and its linearised loop sat on the stability boundary with a damping ratio of zero (Table~\ref{tab:loop}). A uniform scale of all gains produced a stable loop only between about 1.1 and 2.5 times, where the damping ratio never exceeded 0.32. The gain grid scaled the output in this way, although its best gain still failed 170 of 512 paths. A uniform scale stabilised S1 but did not damp it; therefore, a repair had to change the balance of the gains.

\begin{figure*}[pos=tbp]
\centering
\includegraphics{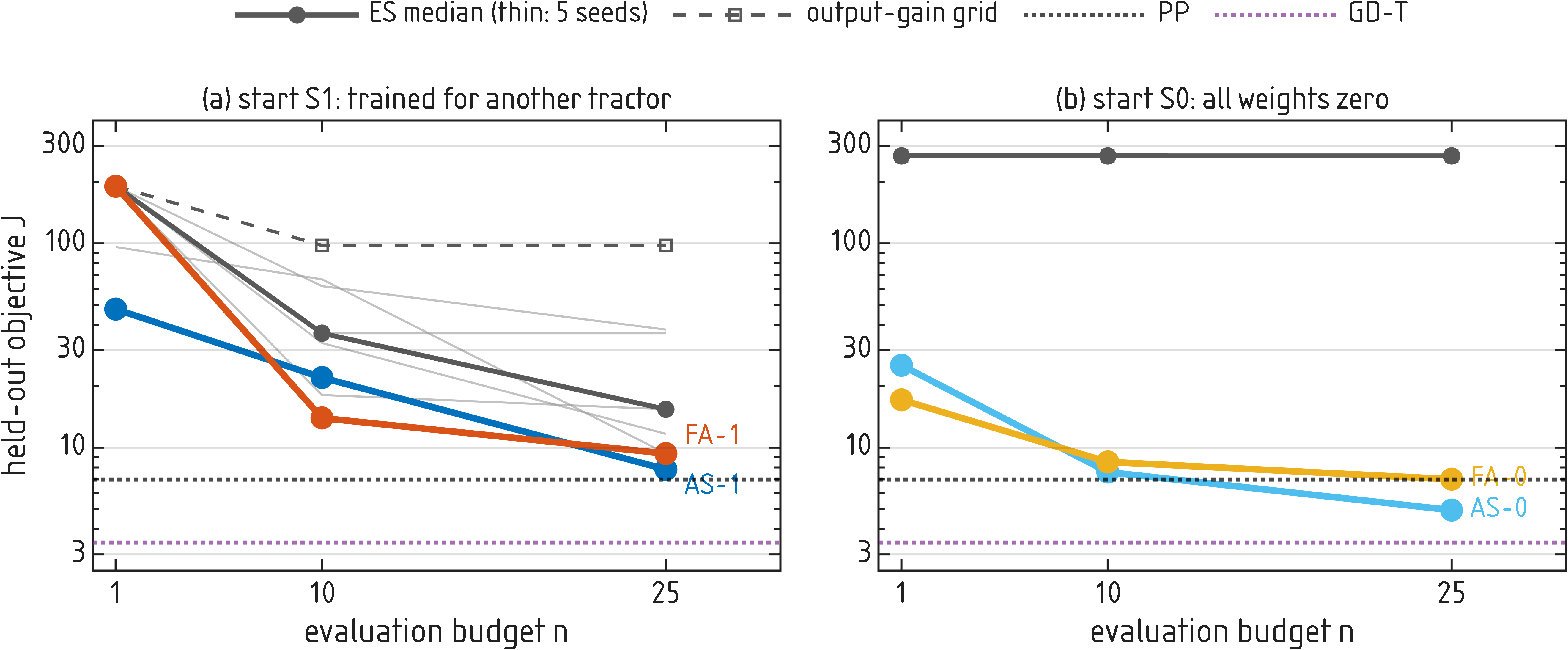}
\caption{Held-out objective $J$ on 512 unseen paths versus the number of simulated evaluations used at 1, 10 and 25, where the left panel shows the repair of S1 and the right the synthesis from zero (S0), comparing the two LLMs against the evolution strategy and the gain grid, while pure pursuit and GD-T appear as horizontal lines; for Fable on S1 the value at 1 evaluation is the start because its first turn was invalid, and on S0 the evolution strategy and the gain grid coincide at about 268.}
\label{fig:anytime}
\end{figure*}

GPT-6 Astra ended with no failures and had few failures already within 10 proposals, while Claude Fable 5.1 had 1 failure in 512 paths; its first turn produced no proposal because the reasoning used up the output limit. Both repairs ended with a lower error than the gain grid and the median of the evolution strategy with the same evaluation budget (Fig.~\ref{fig:anytime}, Table~\ref{tab:heldout}), whereas the evolution strategy came close only with 1000 evaluations. Both repairs first moved the loop in the same direction by reducing the lateral gain and strengthening the inner loop on the command angle, which allowed the first valid proposal of each model to stabilise the linear loop in one edit. With 25 evaluations, both LLMs turned an unstable controller into a working one.

The two sessions used the same prompt, the same text-only access, and the same feedback, and the prompt left the extent of the edit open by stating, ``You may change any subset of the weights: a few, or all of them''. GPT-6 Astra edited the trained network by applying multipliers on groups of weights, which resulted in 144 of 253 weights changing and 32 being set to zero, including all weights on the command history, so that 220 non-zero weights remained at the end; its first proposal stated it ``Retained the learned preview structure''. In contrast, Claude Fable 5.1 deleted the trained network and wrote its own controller into the 253 slots from its first valid proposal, where 227 trained weights were set to zero and none of the others kept its value, leaving 25 non-zero weights at the end; it described this as a ``complete hand redesign as a cascaded controller''. This controller consisted of an inner loop on the command angle, an outer loop on the lateral and heading error, and a curvature feed-forward with preview, which is the same structure as its design from zero. Fig.~\ref{fig:weights} shows both outcomes, and the non-zero weights appear in Table~\ref{tab:heldout}, where 252 existed in the start, 220 for Astra versus 25 for Fable. The two models behaved differently in these sessions.

\subsection{Writing a controller from zero weights}
\label{sec:res-s0}

Both LLMs wrote a working steering controller into the all-zero network, where no failures occurred on the unseen paths (RQ5). The search with the same evaluation budget and the gain grid never left zero output, and the evolution strategy with 1000 evaluations also failed every path (Fig.~\ref{fig:anytime}). On the zero network, a small random change of all weights reaches the output from the path tokens only through 4 weight layers in a row, so that the change of the output is a product of 4 small changes. The evolution strategy started from the noisy zero networks of PPO, with noise of 0.01 or 0.05, and also failed all 512 paths after 1000 evaluations (Table~\ref{tab:heldout}). The first proposal of Fable already worked without a failure, whereas the first proposal of Astra failed 14 paths. After 25 evaluations, the final controller of Astra had a lower error than tuned pure pursuit, and that of Fable had about the same error (Table~\ref{tab:heldout}).

\begin{figure*}[pos=tbp]
\centering
\includegraphics{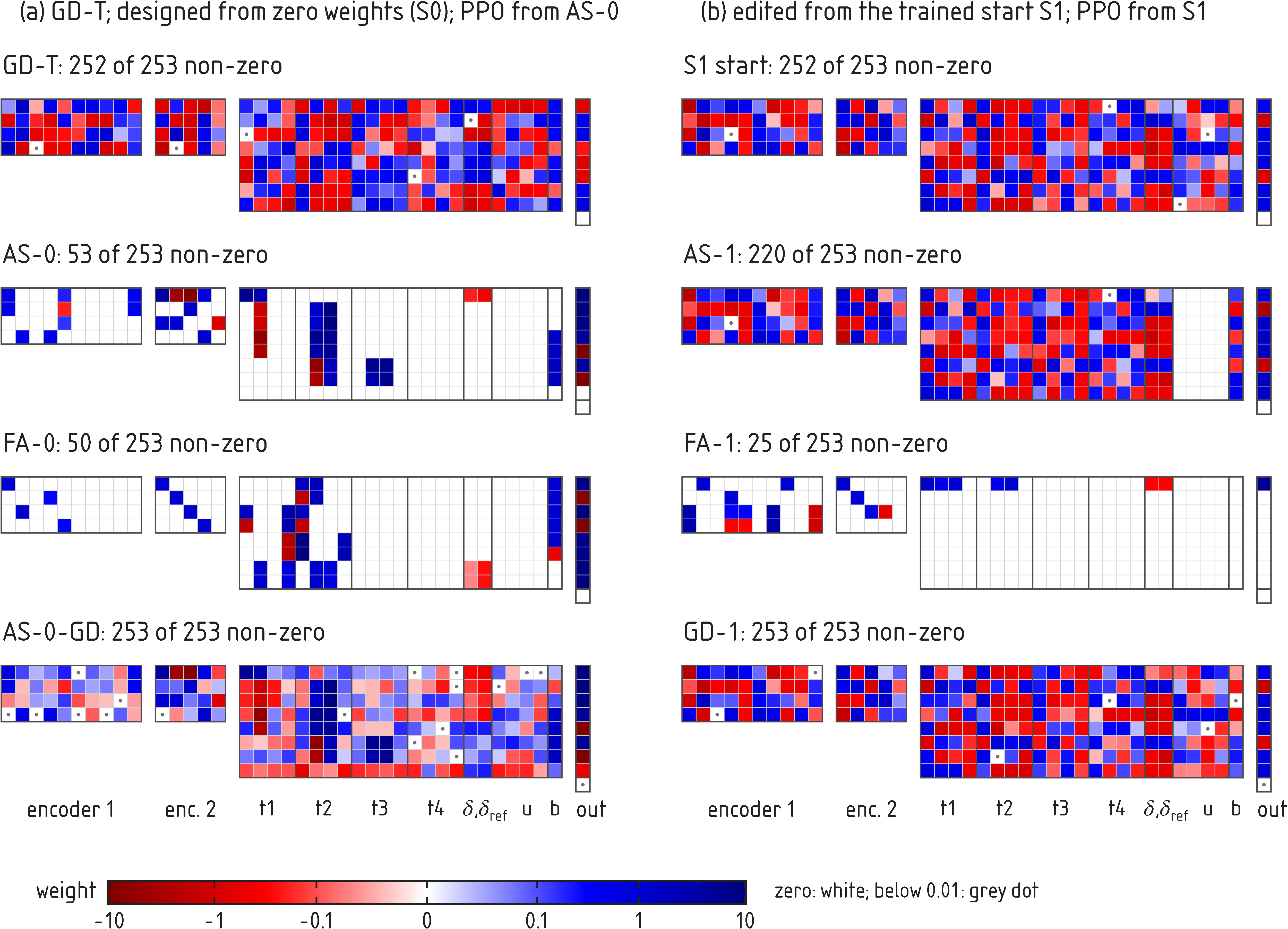}
\caption{Weight maps of all 253 weights of eight networks, showing (a) designs from zero of Astra and Fable, PPO started from the design of Astra (AS-0-GD) and the network trained from scratch by gradient descent (GD-T), and (b) the start S1, its repairs by Astra and Fable and PPO started from S1 (GD-1), where red indicates negative and blue positive values on a logarithmic scale from 0.01 to 10, with pure colours at 0.3 and darker shades towards 10; zero weights are white, non-zero weights below 0.01 are white with a grey dot, and each block title gives its number of non-zero weights.}
\label{fig:weights}
\end{figure*}

GPT-6 Astra and Claude Fable 5.1 produced 53 and 50 non-zero weights of 253 (Fig.~\ref{fig:weights}, Appendix~\ref{app:weights}). Both designs employed a cascaded structure where an inner loop drove the command angle towards a desired angle, which consisted of curvature feed-forward, feedback on the lateral and heading error, and a timed preview of the next curvature. Neither design used the command history. Both designs contained two identical hidden units of the head, namely $g_2$ and $g_3$ for Astra and $g_7$ and $g_8$ for Fable, where each unit had an output weight of 10, the prompt limit, so the pair acted as one unit with a weight of 20. In the Astra design, the fourth output of the token encoder had no weight in the head and did not affect the output. When the tractor remained exactly on an arc, the output of both designs stayed near zero, which indicates that the feed-forward matches the wheelbase of the AMX G-trac.

GPT-6 Astra and Claude Fable 5.1 designed the controllers during the first four turns using long reasoning. The models subsequently tuned the networks by adjusting one physically meaningful parameter at a time, such as the lateral gain, the look-ahead, the lead term or the preview gate, with steady improvement up to the last proposal. GPT-6 Astra provided evidence for this process, stating that it changed only proposal 23's tracking-error weight from 6.8 to 6.0. Both models returned to the best earlier proposal after a worse result occurred. From zero, GPT-6 Astra produced 3 invalid proposals, all of which had a wrong row length, while Claude Fable 5.1 produced 2 invalid proposals, where one weight exceeded the limit and one turn was cut by the output limit (Table~\ref{tab:sessions}).

Both APIs returned summaries of the hidden reasoning (Section~\ref{sec:protocol}), where Fable first performed a long model-based design with phase margin and crossover estimates and placed inner-loop poles by hand. Fable then adjusted one parameter at a time and fitted a parabola through three results to place the lead coefficient; it also named pure pursuit and rejected it because ``it turns early and cuts corners sharply''. Fable diagnosed the recorded worst episode and converted the timing into heading error, stating that ``each 0.1 s of lag costs about 2$^\circ$ of heading''. Fable stated that it left the command history unused because ``they only cover the last 0.2 s and the lag is about 1.2 s'', which are correct values for the network and the total plant delay, and it corrected a wrong model by experiment, noting ``the `fight the compensation' idea was wrong in practice''. Astra first derived the local path as a parabola from the token features and planned pure pursuit with an added damping term on the measured wheel angle, and its later summaries tuned single gains by extrapolating their trend.

\subsection{Comparison with gradient descent}
\label{sec:res-search}

PPO on the same 253 parameters from scratch (GD-T) and fine-tuned from S1 (GD-1) both had a lower error than every LLM controller (Table~\ref{tab:heldout}). PPO from the exact zero network cannot learn, whereas with small random noise on the zero weights, 48 and 38 of 98 seeds ended without a failure at noise levels of 0.01 and 0.05, respectively.

PPO from the design of GPT-6 Astra from zero (AS-0-GD) achieved a lower error than any run of PPO from the noisy zero network (Table~\ref{tab:heldout}), whereas PPO from the repaired S1 networks produced a higher error than PPO from the unedited S1. PPO kept the weights close to the LLM design and filled in the zero weights, which resulted in 190 to 250 weights above 0.01 in magnitude at the end.

\subsection{Mechanism}
\label{sec:res-mechanism}

\begin{figure*}[pos=tbp]
\centering
\includegraphics{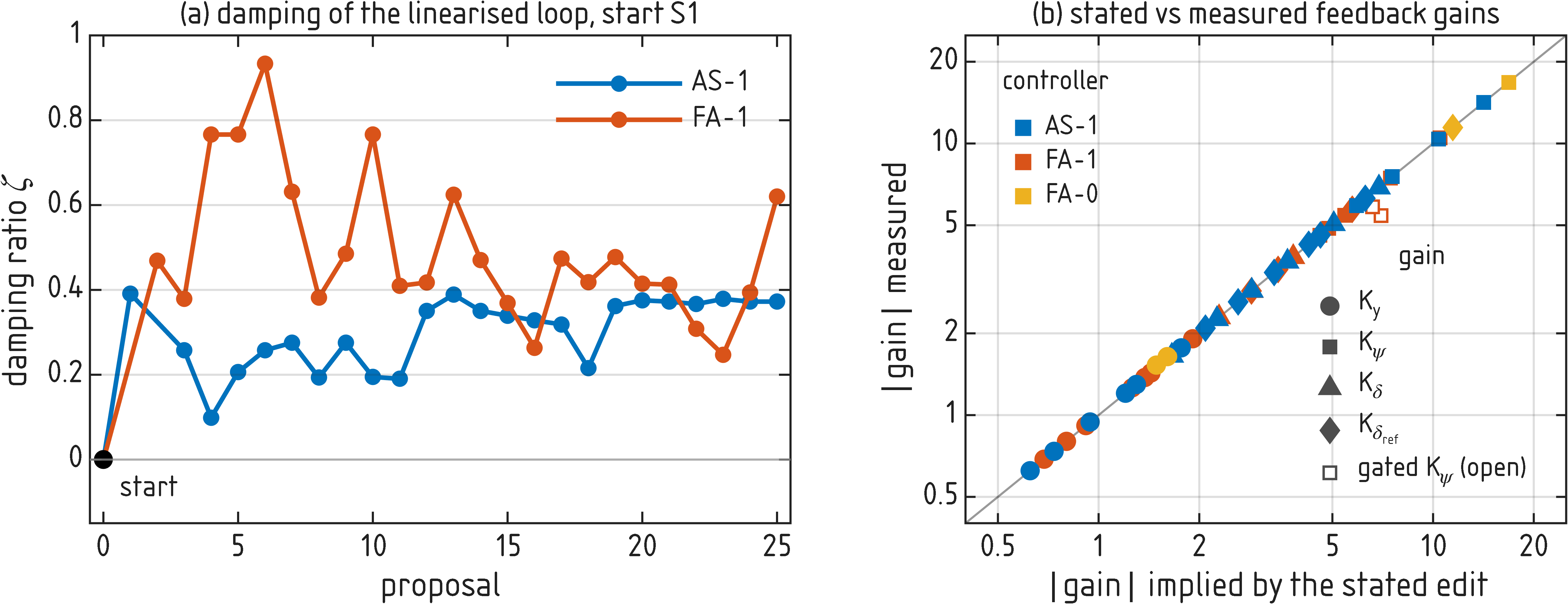}
\caption{(a) Damping ratio of the linearised loop for each proposal of the two repair sessions (S1), and (b) gains implied by the stated edits versus the measured gains for the repairs by Astra and Fable and the design of Fable from zero, where open markers represent the heading gain of Fable's repair gated by the position along the preview.}
\label{fig:mechanism}
\end{figure*}

All final controllers had a stable linearised loop and a damping ratio between 0.26 and 0.63, whereas the start S1 was on the stability boundary (Table~\ref{tab:loop}). All controllers written by the LLMs had zero gain on the command history, while the controllers from gradient descent kept a small one. The designs of both LLMs from zero had about 2 to 2.6 times the heading gain and the inner-loop gain on the command angle of the network trained from scratch (GD-T). Fig.~\ref{fig:mechanism} (a) shows the damping ratio of each proposal in the two repair sessions.

Fable stated its control laws with numbers, and the gains implied by these numbers matched the measured gains within 3~\% in the design from zero, such as the inner-loop gain of turn 1, $-$11.46 stated versus $-$11.459 measured; in the repair session, they matched exactly to two decimals for every inner-loop, lead and lateral gain (Fig.~\ref{fig:mechanism} (b)). One deviation occurred from the middle of the repair session on, where Fable gated part of the heading feedback by the position along the preview, so that the measured heading gain was lower than the stated one. Astra stated its repair edits as multipliers on groups of weights, and every stated multiplier appeared exactly in the measured gains. Astra stated its intents qualitatively and did not predict margins or damping.

\section{Field validation on a real tractor}
\label{sec:field}

\subsection{Test site and path design}
\label{sec:field-path}

This study conducted the field test on 12 November 2026 on a test field of 4.2 ha at the TUM farm in Freising, Germany. The field rises about 17 m towards the south and has a median slope of 4.6 degrees according to a map of the altitude. This study used one test path for all controllers, which was designed on a slope map of the field before the field day. The path avoids slopes of 7 degrees and more, keeps at least 6 m between the rear axle and the field border, and ensures that neighbouring lanes do not overlap to avoid soil compaction. The AMX G-trac drove the whole path in one continuous run while the controller changed at fixed points along the path.

\begin{figure*}[pos=tbp]
\centering
\includegraphics{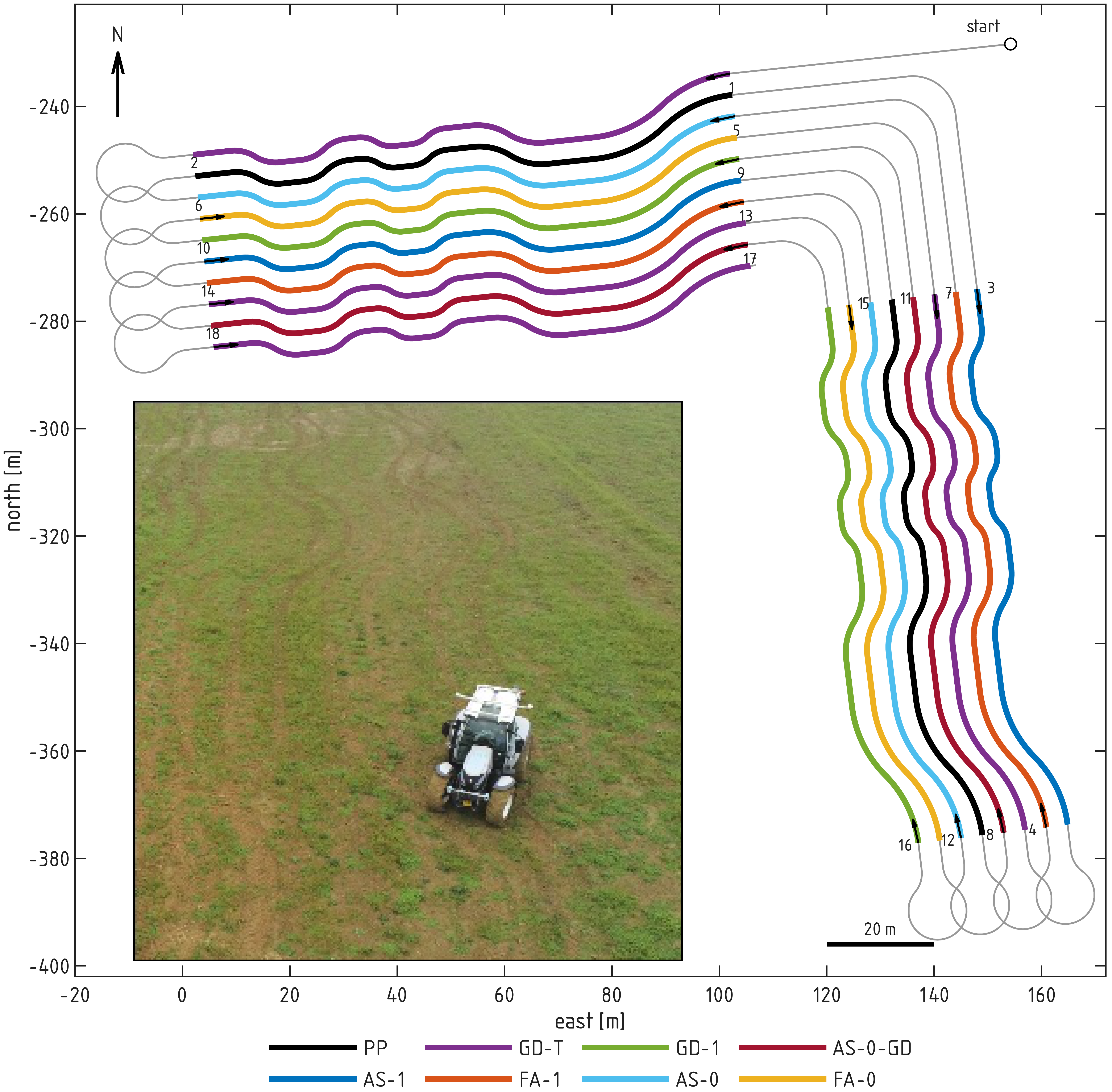}
\caption{The test path of the field run and the lane of each of the 18 blocks coloured by its controller, including the block number, driving direction, start, loop turns and lead-ins in grey, with an inset photo of the AMX G-trac on the test field taken from a drone.}
\label{fig:fieldpath}
\end{figure*}

The path consists of 18 blocks of 112.6 m, where each block comprises a lead-in of 5 m, a lane of 106.6 m and a lead-out of 1 m. The lanes form nested L shapes along the north and east borders of the field, are 4 m apart and were driven alternately in opposite directions, and loop turns join them at the lane ends. Every lane holds the same six S-shaped pairs of arcs, where the arcs of a pair follow each other directly as a reversal from left to right or back. These arcs use curvatures from the same set as the simulated paths (Section~\ref{sec:objective}) up to the largest curvature, with every arc turning 38 degrees and straights of 2 to 9 m between the pairs. A lane driven in the opposite direction has the same curves in reverse order with mirrored signs, and because the controllers are mirror symmetric, every lane gives each controller the same manoeuvres. The whole path is 2779 m long and takes about 46 min at 1 m/s (Fig.~\ref{fig:fieldpath}), and the slope along the path has a median of 3.6 degrees and a maximum of 6.8 degrees.

\subsection{Tractor, controllers and real-time implementation}
\label{sec:field-impl}

The AMX G-trac is a four-wheel-steered research tractor whose front steering is electro-hydraulic and receives the steering angle setpoint over CAN~\citep{Hefele.2024}. In the field test, the rear steering stayed at zero and the speed setpoint was 1.0 m/s, which matched the speed of the simulation plant. RTK GNSS with two antennas provides the position and the heading of the tractor, while the state estimator projects the antenna position to the centre of the rear axle using the measured roll and pitch~\citep{Doering.2024}. A Speedgoat real-time target machine ran the guidance every 50 ms and sent the speed and front steering setpoints over CAN, employing the same setup as in earlier path tracking work on this tractor~\citep{Hefele.2025}.

One Simulink block contained the whole guidance, including the test path, the block schedule and all eight controllers in one MATLAB Function block, which the study generated as real-time code for the Speedgoat. The forward pass of the network was written in MATLAB from the same equations as in Section~\ref{sec:network} and included the 253 weights of each network exactly as written by the LLMs or by the training, with no change. All controllers shared one integrator of the steering-rate command and one history of past commands, so that the steering angle setpoint did not jump when the controller changed.

Eight controllers drove in the field, including pure pursuit, the four controllers written by the LLMs, and three networks trained with PPO (Sections~\ref{sec:baselines} and \ref{sec:gd-from-llm}), and the start network S1 was not driven. Pure pursuit kept its look-ahead of 4.0 m from the simulation and was not tuned again in the field. The Simulink block switched the controller automatically at the start of each block within the lead-in (Fig.~\ref{fig:fieldpath}), and an operator in the cab supervised the run.

\subsection{Evaluation of the field run}
\label{sec:field-eval}

The study measured the lateral error as the cross-track error of the estimated centre of the rear axle to the path, as the controller saw it, only within the lane of each block without the lead-in and the lead-out. The lanes ran alternately in opposite directions across the side slope of the field, and this study calculated the mean RMS lateral error of the two lanes for each controller. The comparison used blocks 2 to 17, where each controller drove one lane along the north side and another along the east side. GD-T also drove blocks 1 and 18 as a warm-up and to check for condition changes, and the comparison excluded them.

\subsection{Field results}
\label{sec:field-results}

\begin{figure}[pos=htbp]
\centering\setlength{\tabcolsep}{3.5pt}
\includegraphics[width=\columnwidth]{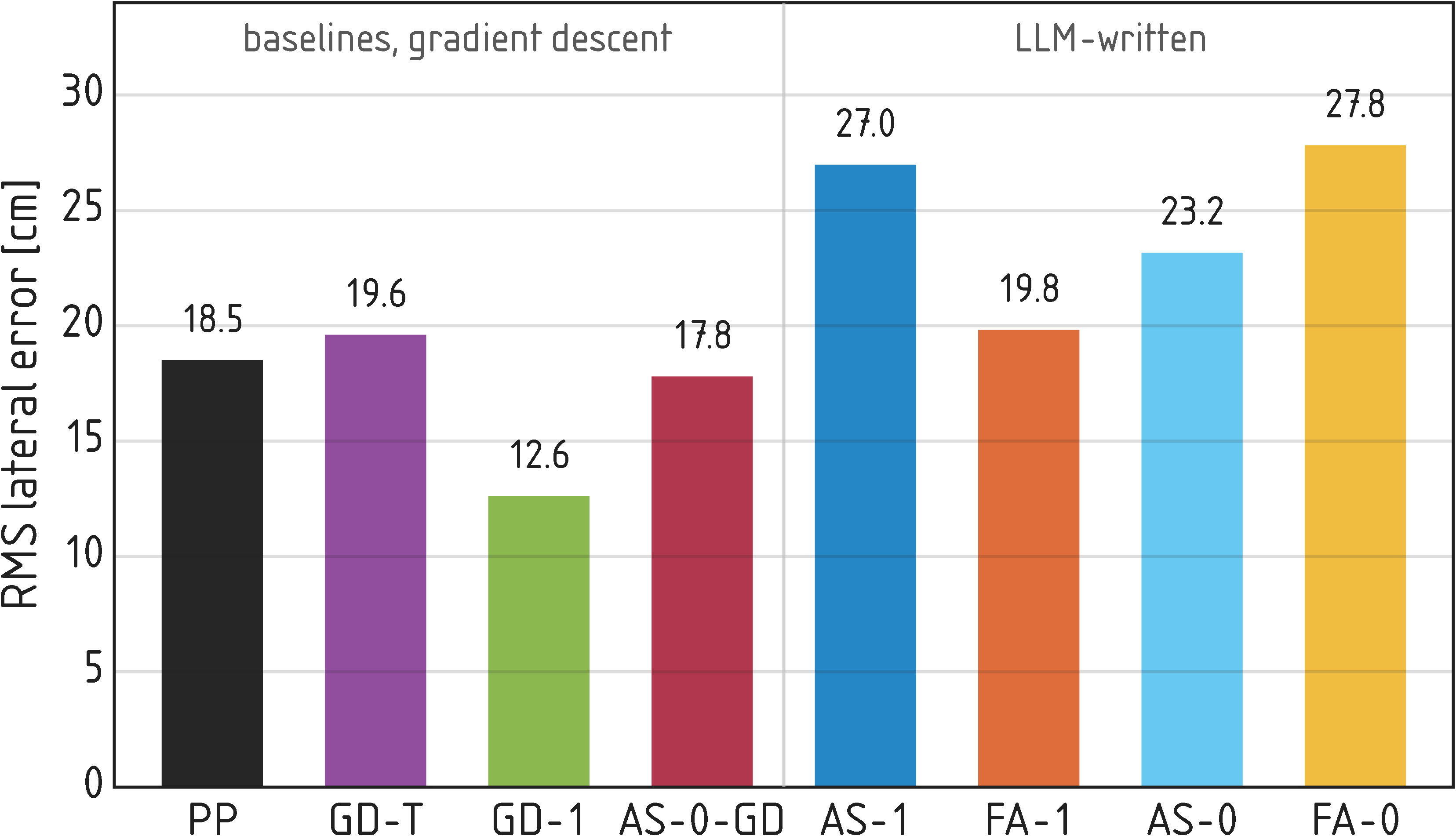}
\caption{RMS lateral error per controller on the real AMX G-trac for comparison blocks 2 to 17, where each bar represents the mean RMS of its two lanes, labelled with its value, showing pure pursuit and networks trained with PPO on the left and controllers written by the LLMs on the right.}
\label{fig:fieldres}
\end{figure}

The AMX G-trac drove the whole test path in one run of 51 min without intervention, and the median speed was 0.93 m/s at the setpoint of 1.0 m/s. All eight controllers steered the tractor through their lanes, including the two networks that the LLMs wrote from all-zero weights (AS-0, FA-0). The RMS lateral error ranged from 12.6 cm (GD-1) to 27.8 cm (FA-0), as shown in Fig.~\ref{fig:fieldres}. The pure pursuit controller achieved 18.5 cm, and the repair by Fable (FA-1) reached 19.8 cm, which was the lowest of the four controllers written by the LLMs. Every controller exhibited the same shape of the error along its two lanes, and all errors peaked at the same positions in the lane. The error of a lane depended on its driving direction across the side slope.

\section{Discussion}
\label{sec:discussion}

\subsection{Engineering rather than search}
\label{sec:disc-engineering}

The designs of the LLMs are sparse, cascaded, and in classical form, with a few active weights per signal path and readable gains. The LLM seems to bring the prior of a control engineer and writes what an engineer would write bottom-up, whereas gradient descent adjusts all weights at once top-down. Even without the control-theory framing (Section~\ref{sec:res-zeroshot}), all ten replies diagnosed the failure in control terms without being asked, and this study makes no claim that the framing caused a difference. Reading a network has been described as local arithmetic plus prior knowledge~\citep{Ahmed.2026}, which is consistent with the reading tests of this study that asked for gains computed by hand. The reasoning summaries show an engineer's workflow where a design from a model comes first, followed by experiments with one parameter at a time and a wrong model corrected by the simulated feedback (Section~\ref{sec:res-s0}). The provider writes these summaries with an undisclosed method, from one session per model, so they show what was documented rather than how the model computed.

The control theory given in the prompt and the linearised loop used for verification made the edits checkable in this study. In this study the LLM acted as a designer that a control engineer can audit, not as a search method.

\subsection{Writing from zero versus repairing}
\label{sec:disc-s0-s1}

Writing from zero (S0) gave better controllers in simulation than repairing (S1) for both LLMs, yielding lower error on the unseen paths and no failures (Table~\ref{tab:heldout}). Both LLMs also reached a working controller with fewer evaluations from zero, where Fable succeeded with its first proposal and Astra within ten. However, the sessions from zero were not cheaper or faster, as both took longer and used more output tokens than the repair sessions (Table~\ref{tab:sessions}). A possible reason is that a dense trained network is hard to read, whereas a blank network lets the LLM impose its own structure. Fable's repair is consistent with this, as it set 227 weights of the trained network to zero and wrote its own controller into it, much like its design from zero (Section~\ref{sec:res-s1}). This study used one session per case and did not test the reason.

\begin{table*}[pos=tbp]
\centering
\def\baselinestretch{1}\scriptsize
\caption{Elements of Weight-Space Engineering as defined in Section~\ref{sec:intro-wse}, compared with weight-space programming, and how this study followed them.}
\label{tab:wse}
\begin{tabular}{@{}>{\raggedright\arraybackslash}p{0.12\textwidth}>{\raggedright\arraybackslash}p{0.21\textwidth}>{\raggedright\arraybackslash}p{0.30\textwidth}>{\raggedright\arraybackslash}p{0.35\textwidth}@{}}
\toprule
Element & Weight-space programming & Weight-Space Engineering & This study \\
\midrule
Core act & An LLM reads, explains, repairs or writes weights directly & The same act, as one step of a process & The LLMs read the network, repaired S1 and wrote S0, without gradients or code execution \\
\addlinespace
Domain & Any task: games, text, images, data & A technical system with physical consequences, e.g. a feedback controller & The steering loop of a tractor with actuator dead time and lag \\
\addlinespace
Input from the human & A task & A specification and the knowledge of the field; not the meaning of single weights & Plant, objective and limits in the prompt; control theory as the field knowledge; no hint on the direction of the edit \\
\addlinespace
Cost & Rarely stated & Explicit costs (evaluations, tokens, time) against the alternatives & Evaluation budget of 25 per session; tokens and time per session (Table~\ref{tab:sessions}); compared with equal-budget search and with PPO \\
\addlinespace
Acceptance and lifecycle & A working demonstration & Verification and validation with the theory of the field, a test on the machine, deployment and maintenance, traceable reasoning & Unseen test paths; exact linearised closed loop; stated control law checked against measured gains; retrofit to a slower actuator (S1); test on the real tractor (Section~\ref{sec:field}) \\
\bottomrule
\end{tabular}
\end{table*}

\subsection{Language models and gradient descent}
\label{sec:disc-gd}

This study interprets the roles of the language model and gradient descent as complementary rather than competitive. The LLM acts as the designer that finds a working structure, while gradient descent acts as the tuner that refines all weights. Gradient descent converged when it started from the designs of the LLMs (Section~\ref{sec:res-search}). Combining the two represents a combination of eras, not a new era (Table~\ref{tab:eras}). A few active weights, coarse values and a small evaluation budget limit the accuracy of the LLM designs.

\subsection{The field test}
\label{sec:disc-field}

All four controllers written by the LLMs, including two repairs and two designs from zero, steered the real tractor along the whole test path without intervention (Section~\ref{sec:field}), and the study used the weights exactly as written without manual correction. The repair by Fable came within about 1 cm of pure pursuit and the weakest LLM controller within about 9 cm.

All controllers showed the same error peaks along the lane and drifted sideways with the side slope, which suggests that the side slope and the tractor, not the controller, set most of the error. The frozen network has no memory and no integral action, and pure pursuit has neither; therefore, both leave a steady offset on a side slope. Accurate path tracking on sloping fields appears to need a compensating element in the controller, such as an integrator or a side slope term from the measured roll. The architecture was fixed before the study for the simulated plant, and the study never set such robustness as the target, so the structure of the network would likely differ if that target were set. The field accuracy therefore does not rank the LLMs as guidance designers; instead, the field test shows that the weights written by the LLMs transfer from simulation to the real tractor (Section~\ref{sec:intro-objectives}).

\subsection{Practical implications and the cost of Weight-Space Engineering}
\label{sec:disc-practice}

Weight-Space Engineering allows the retrofitting of a trained controller to a new machine without retraining, such as one with a slower steering actuator. Working in the weights instead of writing a symbolic control law pays off where a neural controller already exists and the network can be repaired in place, as GPT-6 Astra kept the learnt preview, and a readable design is a good start for gradient descent (Section~\ref{sec:res-search}). The cost of a session comprises tokens, time and the engineer's effort (Table~\ref{tab:sessions}). In contrast, training with proximal policy optimisation requires samples, compute and tuning, which needed 20 million simulated steps per network in this study. Readable controllers with few active weights could possibly provide a basis for safety arguments and certification that a dense network does not provide. Table~\ref{tab:wse} shows that the Weight-Space Engineering process was followed, whereas it does not show that the approach pays off in general.

\subsection{Fable 5.2}
\label{sec:disc-fable52}

Fable 5.2 was released while this paper was being prepared, and this study ran a quick check with it. Fable 5.2 wrote two proposals per task, where the second proposal came after one simulated evaluation. It received the same prompt as Fable 5.1 except for the output instruction. Writing from zero worked in one turn without failures on the unseen paths, where the error of Fable 5.2 was 9.5 cm whereas the first proposal of Fable 5.1 was 17.2 cm, and its stated damping of about 0.8 matched the linearised loop damping of 0.85. The first repair proposal stabilised the linear loop but left failures on the unseen paths, as Fable 5.1 did in one turn. After one simulated evaluation, its second proposal completed all unseen paths, and its error was about the same as that of Fable 5.1, at 16.8 cm and 16.6 cm, respectively. In this quick check, Fable 5.2 succeeded and failed in the same tasks as Fable 5.1 at the same evaluation budget and read the network equally well, and its lower error from zero came from one sample. It differed mainly in reporting, as it stated its uncertainty and how the next evaluation would test its change. In the reading test, Fable 5.2 read the same control law, and 98~\% (90 of 92) of the signs it answered were correct, whereas Fable 5.1 achieved 93~\%. Fable 5.2 also abstained where a sign was uncertain.

\subsection{Strategies of the models}
\label{sec:disc-strategies}

The models seem to choose different strategies on the trained network given the same prompt, access and feedback, where GPT-6 Astra kept and edited the trained network whereas Claude Fable 5.1 deleted it and wrote its own controller, although both repairs worked (Section~\ref{sec:res-s1}). Both models wrote the same kind of controller, a sparse cascade, from zero (Section~\ref{sec:res-s0}), which suggests that the strategies differ where there is something to keep. This resembles two choices in engineering: tuning the existing controller versus redesigning it from first principles. Replacing loses what the training had learnt, such as the learnt preview, but gives a readable controller with few weights, while editing keeps the learnt structure but leaves a dense network that remains hard to read. Fable 5.2 edited the trained network by multipliers like Astra, with no weight set to zero, so the strategy differed between two versions of the same model family and seems to belong to the session or version rather than to the vendor. In the repair of trained game policies, GPT-6 Astra rebuilt the policy and Claude Opus 5.5 made small edits (Section~\ref{sec:rw-wsp}), whereas the two vendors reversed their roles in this study. For Weight-Space Engineering, the choice between keeping and replacing belongs to the specification; if the engineer does not state it, the model makes it.

\subsection{Limitations}
\label{sec:disc-limitations}

This study conducted one session per model and start point without performing statistical tests; furthermore, the repair sessions were the sessions in which the prompt was developed. The prompt of the zero-shot test (RQ1) omitted the control-theory framing, whereas the system prompt of its agents still asked them to reason analytically about what the weights do. Exact reproduction depends on the vendors keeping the two closed, versioned models available. The reading and control-theory tests used a pilot variant of the plant and an earlier network trained on the nominal plant. The simulation plant comes from system identification, and the study covered one tractor, one speed, one test path and one field. The transfer to larger networks remains open since the study used one small network with 253 parameters and an evaluation budget of 25 evaluations. By design, the study compared the controllers only with pure pursuit as the classic guidance baseline, not with nonlinear model predictive control or other state-of-the-art guidance controllers (Section~\ref{sec:intro-objectives}).

\section{Conclusions}
\label{sec:conclusions}

Zero-shot repair without simulation and without the control-theory framing is possible but not reliable: two of ten repairs removed all failures; all ten diagnosed the failure in control terms. Both LLMs read the network correctly, although Astra declined to judge the closed-loop effect of single weights locally. Both LLMs analysed the closed loop correctly on four plants regarding stability, stable gain ranges and limit cycles, where Astra was more precise.

Both LLMs turned an unstable trained network into a working controller with 25 simulated evaluations and outperformed gradient-free search that only approached similar results with 40 times the evaluation budget. Both LLMs wrote sparse cascaded control laws into the all-zero network where search never started to steer; Astra's design outperformed tuned pure pursuit, and Fable's stated gains matched the measured gains. Gradient descent remains more accurate, but a design by an LLM is a good starting point, so the new approach complements training rather than replacing it.

All four controllers written by the LLMs steered the real AMX G-trac without intervention (Section~\ref{sec:field}). This study followed the Weight-Space Engineering process defined in the introduction, element by element (Table~\ref{tab:wse}).

This study presents the first Era-4 controller of tractor guidance (Table~\ref{tab:eras}), which an LLM wrote as network weights without gradients; although its structure differs from pure pursuit, it already tracked the path on the real AMX G-trac at the level of that classic Era-1 controller. Era 4 of tractor guidance starts with this study.

However, the method is not specific to tractors, as Weight-Space Engineering lets an LLM design a controller in the weights of a neural network with control theory as the field knowledge, verified like any engineered controller. This study, as the first instance of Weight-Space Engineering, marks the beginning of a new way of designing controllers in control engineering.

\begingroup\small
\section*{Declaration of generative AI and AI-assisted technologies in the manuscript preparation process}
Large language models are the object of this study: GPT-6 Astra (OpenAI, \texttt{gpt-6-astra}) and Claude Fable 5.1 and 5.2 (Anthropic) wrote the controller weights evaluated in this work; their access and working conditions are described in Sections~\ref{sec:llms} and~\ref{sec:understanding-methods}. During the preparation of this work the author used Claude Opus 5.5 (Anthropic) in order to support coding, data analysis and the literature search, to orchestrate the training runs and to assist with text. After using these tools, the author reviewed and edited the content as needed and takes full responsibility for the content of the published article.

\section*{Acknowledgements}
The author thanks Ruben Hefele for his support in the field test drives. The author also thanks VDMA Landtechnik for its long-term support of our research since 2019.

\par\endgroup
\clearpage
\appendix
\section*{Appendices}
\renewcommand{\theHtable}{\thesection.\arabic{table}}
\renewcommand{\thetable}{\thesection.\arabic{table}}

\section{LLM session protocol and prompt}
\label{app:prompt}
\setcounter{table}{0}

This appendix provides the messages of a session, the verbatim prompt, excerpts of the weight file and one feedback message. GPT-6 Astra and Claude Fable 5.1 worked over their APIs without tools and without a system prompt, and both models received identical message texts in every turn. Each session included up to 25 proposals, where simulation on the 64 feedback paths and a feedback message to the model followed each proposal; the evaluation budget and the controller selection are described in Section~\ref{sec:protocol}. The prompt and the feedback messages call the feedback paths ``training paths'', where both terms mean the same feedback set.

All messages were plain text with Markdown markup. They are reproduced here character for character, with long lines wrapped at the margin (marked \textcolor{gray}{$\hookrightarrow$}); text in italics and square brackets is an editorial placeholder, not part of a message.

\subsection{Messages of a session}
\label{app:prompt-messages}

Turn 1 of a session:

\begin{lstlisting}[style=md]
You take part in a research experiment. The task description and the input files follow, each between lines
'===== BEGIN <name> =====' and '===== END <name> ====='.

===== BEGIN briefing.md =====
(*\textit{[the prompt, Appendix~\ref{app:prompt-text}]}*)
===== END briefing.md =====

===== BEGIN weights.md =====
(*\textit{[the start weights as labelled tables, Appendix~\ref{app:prompt-weights}]}*)
===== END weights.md =====

===== BEGIN weights.json =====
(*\textit{[the same numbers in JSON]}*)
===== END weights.json =====

===== BEGIN metrics.md =====
(*\textit{[the metrics of the start weights, in the format of Appendix~\ref{app:prompt-feedback}]}*)
===== END metrics.md =====

Write proposal 1.
\end{lstlisting}

Each later turn n:

\begin{lstlisting}[style=md]
Feedback for proposal (*\textit{[n\,$-$\,1]}*):

(*\textit{[feedback on proposal n\,$-$\,1, Appendix~\ref{app:prompt-feedback}]}*)
Write proposal (*\textit{[n]}*).
\end{lstlisting}

\subsection{Prompt}
\label{app:prompt-text}

The prompt (\texttt{briefing.md}) of the S1 sessions is reproduced on page~\pageref{app:prompt-box}; Astra and Fable received the same prompt.

\begin{figure*}[pos=tbp]
\phantomsection\label{app:prompt-box}
\begin{lstlisting}[style=md,basicstyle=\def\baselinestretch{1}\fontsize{6.4}{7.2}\selectfont\ttfamily,multicols=2,breakindent=8pt,xleftmargin=0pt,xrightmargin=0pt]
# Task: edit the weights of a tiny steering neural network by hand

You are an expert in control engineering and neural networks. You will **edit the numbers of a small neural network directly**, so that it steers a tractor along a path well. You will get feedback after each proposal (see section 6).

## 0. Working conditions
- This is a text-only exchange. You cannot execute code, run simulations or access anything beyond the text in this conversation.
- There is no training, no gradients and no optimizer.
- All analysis is done by your own reasoning, written out as you see fit. The only output that counts is the set of weight values you propose.

## 1. The target tractor (plant)
- **Vehicle:** a tractor following a planned path of straight lines and circular arcs. It is a kinematic bicycle model with the reference point on the rear axle.
  - Wheelbase L = 2.42 m, constant speed v = 1 m/s, time step dt = 0.05 s.
  - dx/dt = v cos(psi), dy/dt = v sin(psi), dpsi/dt = v tan(delta) / L, where psi is the heading and delta the front wheel angle.
- **Control output:** every 0.05 s the network outputs u in [-1, 1], a steering-**rate** command. Positive u turns left.
- **Steering chain:**
  1. The command is integrated in the controller: delta_ref <- clip(delta_ref + u * 15 deg/s * 0.05 s, -30 deg, +30 deg).
  2. Pure dead time T_d = 0.45 s.
  3. First-order lag with time constant tau = 0.75 s.
  4. The physical wheel limits are +/-34 deg and +/-26.5 deg/s.
  The actual wheel angle delta is measured.
- **Paths:**
  - Each path is 60 m long with 10 segments, and starts at (0, 0) heading along +x.
  - Segment 0 is straight. The other segments are lines or arcs, with curvature from {0, +/-0.25, +/-0.5, +/-0.75, +/-1} * 0.1848 1/m (minimum radius 5.4 m).
  - Curvature jumps at segment junctions; full left-to-right reversals are the hardest case.
  - Before the network sees the path, each segment is split into equal "chunks" of at most 4 m. The network sees the current chunk and the next 3 chunks.
- **An episode fails** when the lateral error |e_y| exceeds 2 m.

## 2. The starting network
The network was trained with reinforcement learning (PPO) on a **different tractor**: wheelbase 2.0 m and a faster steering actuator (dead time T_d = 0.20 s, lag tau = 0.20 s), with the same command integrator and paths. There it tracked the paths well. On the target tractor it performs badly (see `metrics.md`).

**Your job:** change the weights so that the network tracks the paths well on the target tractor of section 1: small lateral error and no failures.

## 3. The network (fixed architecture; only the 253 numbers may change)
The forward pass at every time step works as follows.

**Step 1. Fixed SE(2) layer (no parameters).** Each of the K = 4 path chunks j becomes a 9-feature token in the vehicle frame: origin at the rear axle, x forward, y left.
- x0, y0: chunk start point / 10 m
- cos_th0, sin_th0: chunk start heading relative to the vehicle heading
- kappa: chunk curvature / 0.1848 1/m, so +/-1 means the tightest arc
- L: chunk length / 10 m, so at most 0.4
- x1, y1: chunk end point / 10 m
- valid: 1 for a real path chunk, 0 for the straight virtual extension after the path end

Token 1 is the chunk the vehicle is currently on, and tokens 2 to 4 follow it. When the vehicle is exactly on the path, y0 of token 1 is about -e_y/10, where e_y is the lateral error with left positive.

**Step 2. Token encoder, shared by all 4 tokens:**
- h = tanh(W_e0 * token + b_e0), with h a vector of 4 values (`enc.0`)
- z = tanh(W_e2 * h + b_e2), with z a vector of 4 values (`enc.2`)

**Step 3. Head input** (22 values):
- [z(tok1), z(tok2), z(tok3), z(tok4)], 16 values in total
- delta = (measured wheel angle) / 30 deg
- delta_ref = delta_ref / 30 deg
- u[t-1], u[t-2], u[t-3], u[t-4]: the network's own last 4 outputs

**Step 4. Head:**
- g = tanh(W_h0 * input + b_h0), with g a vector of 8 values (`head.0`)
- f = W_h2 * g + b_h2, a scalar (`head.2`)

**Step 5. Exact mirror symmetry:**
- u = tanh( 0.5 * ( f(obs) - f(mirror(obs)) ) )
- mirror(obs) is the same scene reflected left-right. It flips the sign of y0, sin_th0, kappa, y1, delta, delta_ref and all u[t-i].
- Consequences:
  - The policy is exactly odd, so steering left and right is symmetric by construction.
  - The output bias b_h2 cancels and has **no effect**.
  - Even (left-right symmetric) features cancel unless they interact with odd ones through the tanh layers.

## 4. Approach
The network is a steering controller operating in **closed loop** with the plant of section 1. You may analyse it with
control engineering methods, carried out by your own reasoning (section 0), for example: linearization about
straight-line driving to obtain feedback gains on lateral error, heading error, steering angle and the command history;
the closed-loop characteristic equation, poles and damping; gain, phase and delay margins; describing functions for the
tanh and clip nonlinearities; and feed-forward of the path curvature from the preview tokens. Use whatever analysis you
find useful.

## 5. Rules
- Change only the numbers. Keep every key, shape and row/column order. All values must be finite with |w| <= 10.
- You may change any subset of the weights: a few, or all of them.
- Think analytically: gains, time constants, phase lag, what each unit appears to detect.

## 6. Session protocol and scoring
- You may submit **up to 25 proposals**. Each proposal is simulated on the same 64 training paths on the target tractor. You then receive its metrics, one recorded episode around its worst error, and a table of all your proposals so far.
- **Objective** (lower is better): J = rms_cm + 200 * n_fail / n_paths. Here rms_cm is the mean over paths of the per-path RMS lateral error in cm, and n_fail is the number of failed paths.
- At the end, the proposal with the lowest J on the training paths (or the starting network, if none is better) is taken. It is then evaluated on **different paths of the same kind**, so aim for an improvement that generalises, not one tuned to particular paths.
- An invalid proposal (unparsable, wrong keys or shapes, |w| > 10) uses up its slot and gets no score.
- Each proposal is a complete set of weights. You may start from any earlier proposal or from the starting network.
- **Output of each proposal:** return the complete weights as one fenced block ```json proposal.json ... ```
  (exactly the structure of `weights.json`), followed by a short summary of *what* you changed and *why*, in at most
  15 lines. Feedback for proposal N arrives in the next message.

## 7. Data provided
- `weights.md`: the starting weights as labelled tables (rows = output units, columns = inputs, last column = bias).
- `weights.json`: the same numbers in the exact JSON format your proposals must use.
- `metrics.md`: performance of the starting weights on the 64 training paths on the target tractor, plus one recorded episode around the worst error.
\end{lstlisting}
\end{figure*}

In the prompt of the S0 sessions, the first paragraph of section 2 was replaced by:

\begin{lstlisting}[style=md,basicstyle=\def\baselinestretch{1}\ttfamily\scriptsize]
All 253 weights are **zero**: a blank network that always outputs u = 0. It has never been trained.
\end{lstlisting}

\subsection{Weight file}
\label{app:prompt-weights}

The first two of the four tables of \texttt{weights.md} in the S1 sessions (the values of the S1 start network, column S1 of Table~\ref{tab:weights}). The tables \texttt{head.0.weight} and \texttt{head.2.weight} follow in the same layout.

\begin{lstlisting}[style=md,basicstyle=\def\baselinestretch{1}\ttfamily\scriptsize]
### enc.0.weight (rows = h units, last column = bias enc.0.bias)
| unit | x0 | y0 | cos_th0 | sin_th0 | kappa | L | x1 | y1 | valid | bias |
|---|---|---|---|---|---|---|---|---|---|---|
| h1 | -2.051 | 0.7088 | 0.1233 | 0.7771 | 0.3802 | 0.04369 | -1.137 | -0.3794 | -0.2144 | -0.0294 |
| h2 | -0.1013 | -1.834 | -0.3521 | -0.8255 | -0.3175 | 0.8313 | -0.01417 | -0.3847 | -0.481 | 0.2231 |
| h3 | -1.058 | 1.836 | 0.06747 | -0.0002273 | -0.3957 | 0.326 | 0.174 | 0.04793 | -0.1251 | 0.2013 |
| h4 | 0.4728 | -1.106 | -0.03938 | 0.5011 | -0.262 | 1.057 | 0.5479 | -0.5309 | 0.4985 | -0.1178 |

### enc.2.weight (rows = z units, last column = bias enc.2.bias)
| unit | h1 | h2 | h3 | h4 | bias |
|---|---|---|---|---|---|
| z1 | 1.087 | -0.3175 | 1.534 | -0.02735 | 0.06534 |
| z2 | 0.4671 | 0.9245 | -0.8459 | 0.9738 | -0.09384 |
| z3 | -0.9514 | -0.4017 | 0.4068 | 1.287 | 0.07002 |
| z4 | -1.616 | 1.034 | -0.2136 | 0.2013 | 0.1592 |

(*\textit{[tables head.0.weight and head.2.weight]}*)
\end{lstlisting}

\subsection{Feedback message}
\label{app:prompt-feedback}

The feedback on proposal 1 of the S1 session of Astra (\texttt{feedback\_1.md}), with the recorded episode shortened from 48 rows to 6 rows. The file \texttt{metrics.md} of turn 1 had the same format, with the start weights in place of a proposal and without the table of proposals.

\begin{lstlisting}[style=md,basicstyle=\def\baselinestretch{1}\ttfamily\scriptsize]
## Proposal 1 on the target tractor, 64 training paths (60/253 weights changed)

| metric | value | meaning |
|---|---|---|
| rms_cm | 39.5 | mean over paths of per-path RMS e_y [cm] |
| median_rms_cm | 38.3 | median per-path RMS e_y [cm] |
| p95_cm | 79.9 | mean per-path 95th percentile |e_y| [cm] |
| max_cm | 193 | largest |e_y| on any path [cm] |
| median_path_max_cm | 84.1 | median per-path max |e_y| [cm] |
| steady_cm | 42.9 | RMS e_y away from junctions [cm] |
| junction_cm | 36.2 | RMS e_y within 3 m of a curvature change [cm] |
| straight_rms_cm | 17.8 | RMS e_y on straight segments [cm] |
| arc_rms_cm | 44.9 | RMS e_y on arcs [cm] |
| n_fail | 0 | paths that left the 2 m corridor (failed) |
| n_paths | 64 | paths evaluated |
| u_rms | 0.401 | RMS of the steering-rate command u |

### Recorded episode: worst path, around its largest |e_y| (every 0.25 s)
kappa_path = path curvature at the vehicle [1/m]; e_y > 0: vehicle left of path; e_psi = vehicle heading - path heading.

| t_s | s_m | kappa_path | e_y_cm | e_psi_deg | delta_deg | delta_ref_deg | u |
|---|---|---|---|---|---|---|---|
| 11.85 | 11.81 | 0.1848 | 61.5 | -14.61 | -8.17 | 8.01 | 0.993 |
| 12.1 | 12.08 | 0.1848 | 54.4 | -18.13 | -4.82 | 11.73 | 0.989 |
| 12.35 | 12.34 | 0.1848 | 46.0 | -21.2 | -1.36 | 15.42 | 0.981 |
| 12.6 | 12.59 | 0.1848 | 36.4 | -23.8 | 2.16 | 19.06 | 0.964 |
(*\textit{[42 rows omitted]}*)
| 23.35 | 20.87 | 0.1848 | -134.5 | 17.45 | 23.09 | 16.79 | -0.332 |
| 23.6 | 21.06 | 0.1848 | -126.9 | 17.86 | 21.74 | 15.64 | -0.285 |

Objective J = 39.46

## All proposals so far

| proposal | J | rms_cm | n_fail | max_cm | weights changed |
|---|---|---|---|---|---|
| start | 183.40 | | | | 0 |
| 1 | 39.46 | 39.46 | 0 | 193.4 | 60 |

You have 24 proposal(s) left.
\end{lstlisting}

\section{Network weights of the controllers}
\label{app:weights}
\setcounter{table}{0}

Each path token $j = 1, \ldots, 4$ has nine features $\mathbf{p}_j$ in the vehicle frame: $x_0$, $y_0$, $\cos\theta_0$, $\sin\theta_0$, $\kappa$, $L$, $x_1$, $y_1$ and $m$ (valid flag). The forward pass is
\begin{align*}
h_j &= \tanh(A\,\mathbf{p}_j + a), \\ z_j &= \tanh(B h_j + b), \\
g &= \tanh\big(C\,[z_1, \ldots, z_4, \delta/30^\circ, \delta_\mathrm{ref}/30^\circ, \\ &\qquad u_{t-1}, \ldots, u_{t-4}] + c\big), \\ f &= w^\top g + d,
\end{align*}
and the command is $u = \tanh(0.5\,(f(o) - f(\bar o)))$ with the mirrored observation $\bar o$. The output bias $d$ cancels in this difference and therefore receives no gradient. The prompt files (Appendix~\ref{app:prompt}) name these weights and biases \texttt{enc.0} ($A$, $a$), \texttt{enc.2} ($B$, $b$), \texttt{head.0} ($C$, $c$) and \texttt{head.2} ($w$, $d$). All weights are given here to 4 decimals, whereas the prompts gave the start weights to 4 significant digits (Appendix~\ref{app:prompt}). Table~\ref{tab:weights} lists all weights of the six controllers, exact zeros included. Of the gradient-descent controllers, only GD-1, the repair of S1 by PPO, is listed; its output bias of 0.0012 is the noise added to the exactly zero weights of S1 before training (Section~\ref{sec:gd-from-llm}).

\begin{table*}[pos=tbp]
\centering
\def\baselinestretch{1}\tiny
\setlength\tabcolsep{1pt}
\caption{All 253 weights of the six controllers, to 4 decimals. Rows: unit and its input (bias: the bias of the unit); $h$: encoder layer 1, $z_{j,i}$: encoder layer 2 (shared by the four tokens), $g$: head, $f$: output. Exact zeros are given as 0 in grey; 0.0000 is non-zero but below 0.00005.}
\label{tab:weights}
\begin{tabular}[t]{@{}ll*{6}{r}@{}}
 & & \multicolumn{4}{c}{Repair of S1} & \multicolumn{2}{c}{From zero} \\
\cmidrule(lr){3-6}\cmidrule(l){7-8}
Unit & Input & S1 & AS-1 & FA-1 & GD-1 & AS-0 & FA-0 \\
\midrule
$h_{1}$ & $x_0$ & $-$2.0505 & $-$2.051 & \textcolor{gray}{0} & $-$2.1852 & 0.6 & 1 \\
 & $y_0$ & 0.7088 & 0.2481 & 1 & 0.5242 & \textcolor{gray}{0} & \textcolor{gray}{0} \\
 & $\cos\theta_0$ & 0.1233 & 0.1233 & \textcolor{gray}{0} & 0.1541 & \textcolor{gray}{0} & \textcolor{gray}{0} \\
 & $\sin\theta_0$ & 0.7771 & 0.7771 & \textcolor{gray}{0} & 0.7798 & \textcolor{gray}{0} & \textcolor{gray}{0} \\
 & $\kappa$ & 0.3802 & 0.6463 & \textcolor{gray}{0} & 0.3663 & 0.2 & \textcolor{gray}{0} \\
 & $L$ & 0.0437 & 0.0437 & \textcolor{gray}{0} & 0.0682 & \textcolor{gray}{0} & \textcolor{gray}{0} \\
 & $x_1$ & $-$1.1367 & $-$1.137 & \textcolor{gray}{0} & $-$1.1248 & \textcolor{gray}{0} & \textcolor{gray}{0} \\
 & $y_1$ & $-$0.3794 & $-$0.1328 & 1 & $-$0.4166 & \textcolor{gray}{0} & \textcolor{gray}{0} \\
 & $m$ & $-$0.2144 & $-$0.2144 & \textcolor{gray}{0} & $-$0.196 & \textcolor{gray}{0} & \textcolor{gray}{0} \\
 & bias & $-$0.0294 & 0.2256 & \textcolor{gray}{0} & $-$0.0049 & 0.4 & \textcolor{gray}{0} \\
$h_{2}$ & $x_0$ & $-$0.1013 & $-$0.1013 & \textcolor{gray}{0} & $-$0.1028 & 0.6 & \textcolor{gray}{0} \\
 & $y_0$ & $-$1.8341 & $-$0.6419 & \textcolor{gray}{0} & $-$1.8088 & \textcolor{gray}{0} & \textcolor{gray}{0} \\
 & $\cos\theta_0$ & $-$0.3521 & $-$0.3521 & \textcolor{gray}{0} & $-$0.3814 & \textcolor{gray}{0} & \textcolor{gray}{0} \\
 & $\sin\theta_0$ & $-$0.8255 & $-$0.8255 & 1 & $-$0.8454 & \textcolor{gray}{0} & 0.3 \\
 & $\kappa$ & $-$0.3175 & $-$0.5397 & \textcolor{gray}{0} & $-$0.3371 & $-$0.2 & \textcolor{gray}{0} \\
 & $L$ & 0.8313 & 0.8313 & \textcolor{gray}{0} & 0.8001 & \textcolor{gray}{0} & \textcolor{gray}{0} \\
 & $x_1$ & $-$0.0142 & $-$0.0142 & \textcolor{gray}{0} & $-$0.0336 & \textcolor{gray}{0} & \textcolor{gray}{0} \\
 & $y_1$ & $-$0.3847 & $-$0.1346 & \textcolor{gray}{0} & $-$0.4143 & \textcolor{gray}{0} & \textcolor{gray}{0} \\
 & $m$ & $-$0.481 & $-$0.481 & \textcolor{gray}{0} & $-$0.5169 & \textcolor{gray}{0} & \textcolor{gray}{0} \\
 & bias & 0.2231 & 0.2323 & \textcolor{gray}{0} & 0.1856 & 0.4 & \textcolor{gray}{0} \\
$h_{3}$ & $x_0$ & $-$1.0584 & $-$1.058 & 3 & $-$1.0941 & \textcolor{gray}{0} & \textcolor{gray}{0} \\
 & $y_0$ & 1.8355 & 0.6426 & \textcolor{gray}{0} & 1.8178 & \textcolor{gray}{0} & 1 \\
 & $\cos\theta_0$ & 0.0675 & 0.0675 & \textcolor{gray}{0} & 0.0757 & \textcolor{gray}{0} & \textcolor{gray}{0} \\
 & $\sin\theta_0$ & $-$0.0002 & $-$0.0002 & 0.316 & 0.0924 & \textcolor{gray}{0} & \textcolor{gray}{0} \\
 & $\kappa$ & $-$0.3957 & $-$0.6727 & 0.3 & $-$0.4059 & 0.15 & \textcolor{gray}{0} \\
 & $L$ & 0.326 & 0.326 & \textcolor{gray}{0} & 0.3488 & \textcolor{gray}{0} & \textcolor{gray}{0} \\
 & $x_1$ & 0.174 & 0.174 & 3 & 0.1891 & \textcolor{gray}{0} & \textcolor{gray}{0} \\
 & $y_1$ & 0.0479 & 0.0168 & \textcolor{gray}{0} & 0.0846 & \textcolor{gray}{0} & \textcolor{gray}{0} \\
 & $m$ & $-$0.1251 & $-$0.1251 & \textcolor{gray}{0} & $-$0.0621 & \textcolor{gray}{0} & \textcolor{gray}{0} \\
 & bias & 0.2013 & 0.272 & $-$1.05 & 0.2239 & \textcolor{gray}{0} & \textcolor{gray}{0} \\
$h_{4}$ & $x_0$ & 0.4728 & 0.4728 & 3 & 0.5862 & \textcolor{gray}{0} & \textcolor{gray}{0} \\
 & $y_0$ & $-$1.106 & $-$0.3871 & \textcolor{gray}{0} & $-$0.95 & 0.45 & \textcolor{gray}{0} \\
 & $\cos\theta_0$ & $-$0.0394 & $-$0.0394 & \textcolor{gray}{0} & $-$0.0047 & \textcolor{gray}{0} & \textcolor{gray}{0} \\
 & $\sin\theta_0$ & 0.5011 & 0.5011 & $-$0.316 & 0.7521 & 0.5 & \textcolor{gray}{0} \\
 & $\kappa$ & $-$0.262 & $-$0.4454 & $-$0.3 & $-$0.3876 & \textcolor{gray}{0} & 0.3 \\
 & $L$ & 1.0568 & 1.057 & \textcolor{gray}{0} & 1.0306 & \textcolor{gray}{0} & \textcolor{gray}{0} \\
 & $x_1$ & 0.5479 & 0.5479 & 3 & 0.5454 & \textcolor{gray}{0} & \textcolor{gray}{0} \\
 & $y_1$ & $-$0.5309 & $-$0.1858 & \textcolor{gray}{0} & $-$0.547 & \textcolor{gray}{0} & \textcolor{gray}{0} \\
 & $m$ & 0.4985 & 0.4985 & \textcolor{gray}{0} & 0.4923 & \textcolor{gray}{0} & \textcolor{gray}{0} \\
 & bias & $-$0.1178 & $-$0.1995 & $-$1.05 & $-$0.0947 & \textcolor{gray}{0} & \textcolor{gray}{0} \\
\midrule
$z_{j,1}$ & $h_1$ & 1.0873 & 1.087 & 1 & 1.1882 & 4 & 1 \\
 & $h_2$ & $-$0.3175 & $-$0.3175 & \textcolor{gray}{0} & $-$0.3418 & $-$4 & \textcolor{gray}{0} \\
 & $h_3$ & 1.534 & 1.534 & \textcolor{gray}{0} & 1.5435 & $-$9.1268 & \textcolor{gray}{0} \\
 & $h_4$ & $-$0.0273 & $-$0.0273 & \textcolor{gray}{0} & 0.0111 & 0.64 & \textcolor{gray}{0} \\
 & bias & 0.0653 & 0.0653 & \textcolor{gray}{0} & 0.0822 & \textcolor{gray}{0} & \textcolor{gray}{0} \\
$z_{j,2}$ & $h_1$ & 0.4671 & 0.4671 & \textcolor{gray}{0} & 0.4412 & \textcolor{gray}{0} & \textcolor{gray}{0} \\
 & $h_2$ & 0.9245 & 0.9245 & 1 & 0.9466 & \textcolor{gray}{0} & 1 \\
 & $h_3$ & $-$0.8459 & $-$0.8459 & \textcolor{gray}{0} & $-$0.8016 & 1.3333 & \textcolor{gray}{0} \\
 & $h_4$ & 0.9738 & 0.9738 & \textcolor{gray}{0} & 0.9485 & \textcolor{gray}{0} & \textcolor{gray}{0} \\
 & bias & $-$0.0938 & $-$0.0938 & \textcolor{gray}{0} & $-$0.0863 & \textcolor{gray}{0} & \textcolor{gray}{0} \\
$z_{j,3}$ & $h_1$ & $-$0.9514 & $-$0.9514 & \textcolor{gray}{0} & $-$0.6398 & 1 & \textcolor{gray}{0} \\
 & $h_2$ & $-$0.4017 & $-$0.4017 & \textcolor{gray}{0} & $-$0.377 & 1 & \textcolor{gray}{0} \\
 & $h_3$ & 0.4068 & 0.4068 & 0.6 & 0.7134 & \textcolor{gray}{0} & 1 \\
 & $h_4$ & 1.2871 & 1.287 & $-$0.6 & 1.3264 & \textcolor{gray}{0} & \textcolor{gray}{0} \\
 & bias & 0.07 & 0.07 & \textcolor{gray}{0} & 0.0364 & $-$0.7599 & \textcolor{gray}{0} \\
$z_{j,4}$ & $h_1$ & $-$1.6162 & $-$1.616 & \textcolor{gray}{0} & $-$1.5333 & \textcolor{gray}{0} & \textcolor{gray}{0} \\
 & $h_2$ & 1.034 & 1.034 & \textcolor{gray}{0} & 1.049 & \textcolor{gray}{0} & \textcolor{gray}{0} \\
 & $h_3$ & $-$0.2136 & $-$0.2136 & \textcolor{gray}{0} & $-$0.1729 & \textcolor{gray}{0} & \textcolor{gray}{0} \\
 & $h_4$ & 0.2013 & 0.2013 & \textcolor{gray}{0} & 0.1451 & 1 & 1 \\
 & bias & 0.1592 & 0.1592 & \textcolor{gray}{0} & 0.1299 & \textcolor{gray}{0} & \textcolor{gray}{0} \\
\midrule
$g_{1}$ & $z_{1,1}$ & 0.8507 & 0.8507 & 1.43 & 1.1256 & 6 & \textcolor{gray}{0} \\
 & $z_{1,2}$ & $-$0.1655 & $-$0.1655 & 0.461 & $-$0.1636 & 2.473 & \textcolor{gray}{0} \\
 & $z_{1,3}$ & 0.0275 & 0.0276 & 1.046 & 0.0113 & \textcolor{gray}{0} & \textcolor{gray}{0} \\
 & $z_{1,4}$ & $-$0.6297 & $-$0.6297 & \textcolor{gray}{0} & $-$0.684 & \textcolor{gray}{0} & \textcolor{gray}{0} \\
 & $z_{2,1}$ & 0.4336 & 0.2602 & \textcolor{gray}{0} & 0.7673 & \textcolor{gray}{0} & 1.75 \\
 & $z_{2,2}$ & $-$0.7104 & $-$0.4262 & 0.461 & $-$0.6566 & \textcolor{gray}{0} & 2 \\
 & $z_{2,3}$ & $-$0.4299 & $-$0.2579 & 1.046 & $-$0.485 & \textcolor{gray}{0} & \textcolor{gray}{0} \\
 & $z_{2,4}$ & $-$0.4458 & $-$0.2675 & \textcolor{gray}{0} & $-$0.4968 & \textcolor{gray}{0} & \textcolor{gray}{0} \\
 & $z_{3,1}$ & 0.215 & 0.3884 & \textcolor{gray}{0} & 0.1533 & \textcolor{gray}{0} & \textcolor{gray}{0} \\
 & $z_{3,2}$ & 0.2034 & $-$0.0808 & \textcolor{gray}{0} & 0.2858 & \textcolor{gray}{0} & \textcolor{gray}{0} \\
 & $z_{3,3}$ & $-$0.1751 & $-$0.3471 & \textcolor{gray}{0} & $-$0.1887 & \textcolor{gray}{0} & \textcolor{gray}{0} \\
 & $z_{3,4}$ & $-$0.4668 & $-$0.6451 & \textcolor{gray}{0} & $-$0.4438 & \textcolor{gray}{0} & \textcolor{gray}{0} \\
 & $z_{4,1}$ & $-$0.1785 & $-$0.1785 & \textcolor{gray}{0} & $-$0.3431 & \textcolor{gray}{0} & \textcolor{gray}{0} \\
 & $z_{4,2}$ & 0.0079 & 0.0079 & \textcolor{gray}{0} & 0.0465 & \textcolor{gray}{0} & \textcolor{gray}{0} \\
 & $z_{4,3}$ & 0.3615 & 0.3615 & \textcolor{gray}{0} & 0.3452 & \textcolor{gray}{0} & \textcolor{gray}{0} \\
 & $z_{4,4}$ & 0.5749 & 0.5749 & \textcolor{gray}{0} & 0.6417 & \textcolor{gray}{0} & \textcolor{gray}{0} \\
 & $\delta$ & $-$0.0317 & $-$0.0127 & $-$0.3 & $-$0.0443 & $-$0.2 & \textcolor{gray}{0} \\
 & $\delta_\mathrm{ref}$ & 0.0251 & 0.0316 & $-$0.3 & $-$0.1285 & $-$0.4 & \textcolor{gray}{0} \\
 & $u_{t-1}$ & 0.0116 & \textcolor{gray}{0} & \textcolor{gray}{0} & $-$0.0651 & \textcolor{gray}{0} & \textcolor{gray}{0} \\
 & $u_{t-2}$ & $-$0.3224 & \textcolor{gray}{0} & \textcolor{gray}{0} & $-$0.2337 & \textcolor{gray}{0} & \textcolor{gray}{0} \\
 & $u_{t-3}$ & 0.4175 & \textcolor{gray}{0} & \textcolor{gray}{0} & 0.4044 & \textcolor{gray}{0} & \textcolor{gray}{0} \\
 & $u_{t-4}$ & 0.3654 & \textcolor{gray}{0} & \textcolor{gray}{0} & 0.1956 & \textcolor{gray}{0} & \textcolor{gray}{0} \\
 & bias & $-$0.0458 & 0.1406 & \textcolor{gray}{0} & $-$0.0294 & \textcolor{gray}{0} & 1 \\
$g_{2}$ & $z_{1,1}$ & $-$1.2324 & $-$1.232 & \textcolor{gray}{0} & $-$1.3651 & \textcolor{gray}{0} & \textcolor{gray}{0} \\
 & $z_{1,2}$ & 1.093 & 1.093 & \textcolor{gray}{0} & 1.1775 & $-$1.2365 & \textcolor{gray}{0} \\
\bottomrule
\end{tabular}\hfill
\begin{tabular}[t]{@{}ll*{6}{r}@{}}
 & & \multicolumn{4}{c}{Repair of S1} & \multicolumn{2}{c}{From zero} \\
\cmidrule(lr){3-6}\cmidrule(l){7-8}
Unit & Input & S1 & AS-1 & FA-1 & GD-1 & AS-0 & FA-0 \\
\midrule
$g_{2}$ & $z_{1,3}$ & $-$0.0304 & $-$0.0304 & \textcolor{gray}{0} & 0.1966 & \textcolor{gray}{0} & \textcolor{gray}{0} \\
 & $z_{1,4}$ & 1.364 & 1.364 & \textcolor{gray}{0} & 1.3552 & \textcolor{gray}{0} & \textcolor{gray}{0} \\
 & $z_{2,1}$ & $-$0.3552 & $-$0.2131 & \textcolor{gray}{0} & $-$0.6197 & \textcolor{gray}{0} & $-$1.75 \\
 & $z_{2,2}$ & 0.3614 & 0.2168 & \textcolor{gray}{0} & 0.2112 & 1.2365 & 2 \\
 & $z_{2,3}$ & $-$0.3135 & $-$0.1881 & \textcolor{gray}{0} & $-$0.2383 & 10 & \textcolor{gray}{0} \\
 & $z_{2,4}$ & 0.3065 & 0.1839 & \textcolor{gray}{0} & 0.5038 & \textcolor{gray}{0} & \textcolor{gray}{0} \\
 & $z_{3,1}$ & 0.2221 & 0.08 & \textcolor{gray}{0} & 0.2307 & \textcolor{gray}{0} & \textcolor{gray}{0} \\
 & $z_{3,2}$ & $-$0.1208 & 0.0238 & \textcolor{gray}{0} & $-$0.2446 & \textcolor{gray}{0} & \textcolor{gray}{0} \\
 & $z_{3,3}$ & 0.4726 & 0.3472 & \textcolor{gray}{0} & 0.5302 & \textcolor{gray}{0} & \textcolor{gray}{0} \\
 & $z_{3,4}$ & $-$0.3874 & $-$0.2648 & \textcolor{gray}{0} & $-$0.441 & \textcolor{gray}{0} & \textcolor{gray}{0} \\
 & $z_{4,1}$ & 0.114 & 0.114 & \textcolor{gray}{0} & 0.1752 & \textcolor{gray}{0} & \textcolor{gray}{0} \\
 & $z_{4,2}$ & 0.1625 & 0.1625 & \textcolor{gray}{0} & 0.1474 & \textcolor{gray}{0} & \textcolor{gray}{0} \\
 & $z_{4,3}$ & $-$0.1795 & $-$0.1795 & \textcolor{gray}{0} & $-$0.1181 & \textcolor{gray}{0} & \textcolor{gray}{0} \\
 & $z_{4,4}$ & 0.3049 & 0.3049 & \textcolor{gray}{0} & 0.1905 & \textcolor{gray}{0} & \textcolor{gray}{0} \\
 & $\delta$ & 0.9386 & 1.1471 & \textcolor{gray}{0} & 0.9012 & \textcolor{gray}{0} & \textcolor{gray}{0} \\
 & $\delta_\mathrm{ref}$ & 0.6878 & 0.8666 & \textcolor{gray}{0} & 0.7704 & \textcolor{gray}{0} & \textcolor{gray}{0} \\
 & $u_{t-1}$ & 0.0705 & \textcolor{gray}{0} & \textcolor{gray}{0} & 0.02 & \textcolor{gray}{0} & \textcolor{gray}{0} \\
 & $u_{t-2}$ & $-$0.0644 & \textcolor{gray}{0} & \textcolor{gray}{0} & $-$0.2827 & \textcolor{gray}{0} & \textcolor{gray}{0} \\
 & $u_{t-3}$ & $-$0.0113 & \textcolor{gray}{0} & \textcolor{gray}{0} & $-$0.0283 & \textcolor{gray}{0} & \textcolor{gray}{0} \\
 & $u_{t-4}$ & $-$0.1601 & \textcolor{gray}{0} & \textcolor{gray}{0} & 0.0202 & \textcolor{gray}{0} & \textcolor{gray}{0} \\
 & bias & $-$0.3056 & $-$0.4324 & \textcolor{gray}{0} & $-$0.2598 & \textcolor{gray}{0} & 1 \\
$g_{3}$ & $z_{1,1}$ & 0.2991 & 0.2991 & \textcolor{gray}{0} & 0.235 & \textcolor{gray}{0} & 1.75 \\
 & $z_{1,2}$ & 0.1082 & 0.1082 & \textcolor{gray}{0} & 0.2312 & $-$1.2365 & \textcolor{gray}{0} \\
 & $z_{1,3}$ & 0.1177 & 0.1177 & \textcolor{gray}{0} & 0.1184 & \textcolor{gray}{0} & \textcolor{gray}{0} \\
 & $z_{1,4}$ & $-$0.2427 & $-$0.2427 & \textcolor{gray}{0} & $-$0.0849 & \textcolor{gray}{0} & 3.1 \\
 & $z_{2,1}$ & 1.3277 & 0.7968 & \textcolor{gray}{0} & 1.0493 & \textcolor{gray}{0} & 1.75 \\
 & $z_{2,2}$ & $-$0.8058 & $-$0.4835 & \textcolor{gray}{0} & $-$0.695 & 1.2365 & \textcolor{gray}{0} \\
 & $z_{2,3}$ & $-$0.2866 & $-$0.172 & \textcolor{gray}{0} & $-$0.4007 & 10 & \textcolor{gray}{0} \\
 & $z_{2,4}$ & $-$0.4793 & $-$0.2876 & \textcolor{gray}{0} & $-$0.581 & \textcolor{gray}{0} & \textcolor{gray}{0} \\
 & $z_{3,1}$ & 0.2968 & 0.828 & \textcolor{gray}{0} & 0.3503 & \textcolor{gray}{0} & \textcolor{gray}{0} \\
 & $z_{3,2}$ & $-$0.5032 & $-$0.8255 & \textcolor{gray}{0} & $-$0.5278 & \textcolor{gray}{0} & \textcolor{gray}{0} \\
 & $z_{3,3}$ & $-$0.2433 & $-$0.3579 & \textcolor{gray}{0} & $-$0.2558 & \textcolor{gray}{0} & \textcolor{gray}{0} \\
 & $z_{3,4}$ & $-$0.1544 & $-$0.3461 & \textcolor{gray}{0} & $-$0.2623 & \textcolor{gray}{0} & \textcolor{gray}{0} \\
 & $z_{4,1}$ & 0.3432 & 0.3432 & \textcolor{gray}{0} & 0.3671 & \textcolor{gray}{0} & \textcolor{gray}{0} \\
 & $z_{4,2}$ & 0.0685 & 0.0685 & \textcolor{gray}{0} & 0.0093 & \textcolor{gray}{0} & \textcolor{gray}{0} \\
 & $z_{4,3}$ & 0.3816 & 0.3816 & \textcolor{gray}{0} & 0.38 & \textcolor{gray}{0} & \textcolor{gray}{0} \\
 & $z_{4,4}$ & 0.1522 & 0.1522 & \textcolor{gray}{0} & 0.1105 & \textcolor{gray}{0} & \textcolor{gray}{0} \\
 & $\delta$ & $-$0.8402 & $-$0.9811 & \textcolor{gray}{0} & $-$0.9509 & \textcolor{gray}{0} & \textcolor{gray}{0} \\
 & $\delta_\mathrm{ref}$ & $-$0.5309 & $-$0.6689 & \textcolor{gray}{0} & $-$0.6668 & \textcolor{gray}{0} & \textcolor{gray}{0} \\
 & $u_{t-1}$ & 0.3596 & \textcolor{gray}{0} & \textcolor{gray}{0} & 0.3972 & \textcolor{gray}{0} & \textcolor{gray}{0} \\
 & $u_{t-2}$ & $-$0.1489 & \textcolor{gray}{0} & \textcolor{gray}{0} & $-$0.0721 & \textcolor{gray}{0} & \textcolor{gray}{0} \\
 & $u_{t-3}$ & 0.0000 & \textcolor{gray}{0} & \textcolor{gray}{0} & $-$0.1831 & \textcolor{gray}{0} & \textcolor{gray}{0} \\
 & $u_{t-4}$ & 0.1187 & \textcolor{gray}{0} & \textcolor{gray}{0} & 0.0893 & \textcolor{gray}{0} & \textcolor{gray}{0} \\
 & bias & 0.0409 & 0.4431 & \textcolor{gray}{0} & $-$0.0097 & \textcolor{gray}{0} & 1 \\
$g_{4}$ & $z_{1,1}$ & $-$0.0325 & $-$0.0325 & \textcolor{gray}{0} & $-$0.0864 & \textcolor{gray}{0} & $-$1.75 \\
 & $z_{1,2}$ & $-$0.112 & $-$0.112 & \textcolor{gray}{0} & 0.0849 & $-$6 & \textcolor{gray}{0} \\
 & $z_{1,3}$ & $-$0.6711 & $-$0.6711 & \textcolor{gray}{0} & $-$0.6902 & \textcolor{gray}{0} & \textcolor{gray}{0} \\
 & $z_{1,4}$ & 0.0613 & 0.0613 & \textcolor{gray}{0} & 0.2409 & \textcolor{gray}{0} & 3.1 \\
 & $z_{2,1}$ & 0.2927 & 0.1756 & \textcolor{gray}{0} & 0.0607 & \textcolor{gray}{0} & $-$1.75 \\
 & $z_{2,2}$ & $-$0.3848 & $-$0.2309 & \textcolor{gray}{0} & $-$0.3677 & 6 & \textcolor{gray}{0} \\
 & $z_{2,3}$ & $-$0.4236 & $-$0.2542 & \textcolor{gray}{0} & $-$0.5498 & 6 & \textcolor{gray}{0} \\
 & $z_{2,4}$ & $-$0.4953 & $-$0.2972 & \textcolor{gray}{0} & $-$0.5573 & \textcolor{gray}{0} & \textcolor{gray}{0} \\
 & $z_{3,1}$ & 0.3562 & 0.4733 & \textcolor{gray}{0} & 0.3275 & \textcolor{gray}{0} & \textcolor{gray}{0} \\
 & $z_{3,2}$ & 0.0228 & $-$0.1311 & \textcolor{gray}{0} & $-$0.1353 & \textcolor{gray}{0} & \textcolor{gray}{0} \\
 & $z_{3,3}$ & 0.0458 & $-$0.1236 & \textcolor{gray}{0} & $-$0.0362 & \textcolor{gray}{0} & \textcolor{gray}{0} \\
 & $z_{3,4}$ & $-$0.0951 & $-$0.2933 & \textcolor{gray}{0} & $-$0.1939 & \textcolor{gray}{0} & \textcolor{gray}{0} \\
 & $z_{4,1}$ & 0.0158 & 0.0158 & \textcolor{gray}{0} & $-$0.0123 & \textcolor{gray}{0} & \textcolor{gray}{0} \\
 & $z_{4,2}$ & $-$0.5134 & $-$0.5134 & \textcolor{gray}{0} & $-$0.6221 & \textcolor{gray}{0} & \textcolor{gray}{0} \\
 & $z_{4,3}$ & $-$0.3947 & $-$0.3947 & \textcolor{gray}{0} & $-$0.4688 & \textcolor{gray}{0} & \textcolor{gray}{0} \\
 & $z_{4,4}$ & $-$0.4166 & $-$0.4166 & \textcolor{gray}{0} & $-$0.4929 & \textcolor{gray}{0} & \textcolor{gray}{0} \\
 & $\delta$ & $-$1.3381 & $-$1.3184 & \textcolor{gray}{0} & $-$1.5748 & \textcolor{gray}{0} & \textcolor{gray}{0} \\
 & $\delta_\mathrm{ref}$ & $-$0.3937 & $-$0.4961 & \textcolor{gray}{0} & $-$0.6921 & \textcolor{gray}{0} & \textcolor{gray}{0} \\
 & $u_{t-1}$ & 0.7557 & \textcolor{gray}{0} & \textcolor{gray}{0} & 0.7831 & \textcolor{gray}{0} & \textcolor{gray}{0} \\
 & $u_{t-2}$ & 0.5327 & \textcolor{gray}{0} & \textcolor{gray}{0} & 0.5421 & \textcolor{gray}{0} & \textcolor{gray}{0} \\
 & $u_{t-3}$ & 0.5418 & \textcolor{gray}{0} & \textcolor{gray}{0} & 0.4529 & \textcolor{gray}{0} & \textcolor{gray}{0} \\
 & $u_{t-4}$ & 0.6806 & \textcolor{gray}{0} & \textcolor{gray}{0} & 0.6772 & \textcolor{gray}{0} & \textcolor{gray}{0} \\
 & bias & $-$0.2885 & $-$0.1505 & \textcolor{gray}{0} & $-$0.3785 & 1.5 & 1 \\
$g_{5}$ & $z_{1,1}$ & 0.8122 & 0.8122 & \textcolor{gray}{0} & 0.8737 & \textcolor{gray}{0} & \textcolor{gray}{0} \\
 & $z_{1,2}$ & $-$0.9785 & $-$0.9785 & \textcolor{gray}{0} & $-$0.9993 & $-$3 & \textcolor{gray}{0} \\
 & $z_{1,3}$ & $-$0.5089 & $-$0.5089 & \textcolor{gray}{0} & $-$0.6669 & \textcolor{gray}{0} & \textcolor{gray}{0} \\
 & $z_{1,4}$ & $-$0.5464 & $-$0.5464 & \textcolor{gray}{0} & $-$0.519 & \textcolor{gray}{0} & $-$2.3 \\
 & $z_{2,1}$ & 0.4628 & 0.2777 & \textcolor{gray}{0} & 0.875 & \textcolor{gray}{0} & 10 \\
 & $z_{2,2}$ & $-$1.2146 & $-$0.729 & \textcolor{gray}{0} & $-$1.1707 & 3 & \textcolor{gray}{0} \\
 & $z_{2,3}$ & 0.0546 & 0.0328 & \textcolor{gray}{0} & $-$0.0433 & 6 & \textcolor{gray}{0} \\
 & $z_{2,4}$ & $-$0.2011 & $-$0.1207 & \textcolor{gray}{0} & $-$0.2369 & \textcolor{gray}{0} & 2.3 \\
 & $z_{3,1}$ & $-$0.2977 & $-$0.1126 & \textcolor{gray}{0} & $-$0.3322 & \textcolor{gray}{0} & \textcolor{gray}{0} \\
 & $z_{3,2}$ & 0.0645 & $-$0.4215 & \textcolor{gray}{0} & 0.1247 & \textcolor{gray}{0} & \textcolor{gray}{0} \\
 & $z_{3,3}$ & 0.1929 & 0.2147 & \textcolor{gray}{0} & 0.1497 & \textcolor{gray}{0} & \textcolor{gray}{0} \\
 & $z_{3,4}$ & $-$0.417 & $-$0.4974 & \textcolor{gray}{0} & $-$0.3746 & \textcolor{gray}{0} & \textcolor{gray}{0} \\
 & $z_{4,1}$ & 0.2082 & 0.2082 & \textcolor{gray}{0} & 0.0356 & \textcolor{gray}{0} & \textcolor{gray}{0} \\
 & $z_{4,2}$ & 0.2942 & 0.2942 & \textcolor{gray}{0} & 0.2944 & \textcolor{gray}{0} & \textcolor{gray}{0} \\
 & $z_{4,3}$ & 0.4077 & 0.4077 & \textcolor{gray}{0} & 0.3502 & \textcolor{gray}{0} & \textcolor{gray}{0} \\
 & $z_{4,4}$ & $-$0.0395 & $-$0.0395 & \textcolor{gray}{0} & 0.0165 & \textcolor{gray}{0} & \textcolor{gray}{0} \\
 & $\delta$ & $-$0.5159 & $-$0.7109 & \textcolor{gray}{0} & $-$0.5257 & \textcolor{gray}{0} & \textcolor{gray}{0} \\
\bottomrule
\end{tabular}\hfill
\begin{tabular}[t]{@{}ll*{6}{r}@{}}
 & & \multicolumn{4}{c}{Repair of S1} & \multicolumn{2}{c}{From zero} \\
\cmidrule(lr){3-6}\cmidrule(l){7-8}
Unit & Input & S1 & AS-1 & FA-1 & GD-1 & AS-0 & FA-0 \\
\midrule
$g_{5}$ & $\delta_\mathrm{ref}$ & $-$0.527 & $-$0.664 & \textcolor{gray}{0} & $-$0.6445 & \textcolor{gray}{0} & \textcolor{gray}{0} \\
 & $u_{t-1}$ & 0.0852 & \textcolor{gray}{0} & \textcolor{gray}{0} & 0.013 & \textcolor{gray}{0} & \textcolor{gray}{0} \\
 & $u_{t-2}$ & $-$0.0606 & \textcolor{gray}{0} & \textcolor{gray}{0} & 0.0407 & \textcolor{gray}{0} & \textcolor{gray}{0} \\
 & $u_{t-3}$ & 0.0572 & \textcolor{gray}{0} & \textcolor{gray}{0} & 0.0091 & \textcolor{gray}{0} & \textcolor{gray}{0} \\
 & $u_{t-4}$ & $-$0.0178 & \textcolor{gray}{0} & \textcolor{gray}{0} & $-$0.1695 & \textcolor{gray}{0} & \textcolor{gray}{0} \\
 & bias & 0.0259 & 0.2231 & \textcolor{gray}{0} & $-$0.0268 & 1.5 & 1.2 \\
$g_{6}$ & $z_{1,1}$ & $-$0.7417 & $-$0.7417 & \textcolor{gray}{0} & $-$0.7542 & \textcolor{gray}{0} & \textcolor{gray}{0} \\
 & $z_{1,2}$ & 0.3522 & 0.3522 & \textcolor{gray}{0} & 0.2802 & \textcolor{gray}{0} & \textcolor{gray}{0} \\
 & $z_{1,3}$ & 0.1905 & 0.1905 & \textcolor{gray}{0} & 0.1445 & \textcolor{gray}{0} & \textcolor{gray}{0} \\
 & $z_{1,4}$ & 0.0234 & 0.0234 & \textcolor{gray}{0} & $-$0.1741 & \textcolor{gray}{0} & $-$2.3 \\
 & $z_{2,1}$ & $-$0.8393 & $-$0.5036 & \textcolor{gray}{0} & $-$0.6363 & \textcolor{gray}{0} & 10 \\
 & $z_{2,2}$ & 0.9193 & 0.5516 & \textcolor{gray}{0} & 0.865 & $-$6 & \textcolor{gray}{0} \\
 & $z_{2,3}$ & 0.358 & 0.2148 & \textcolor{gray}{0} & 0.3936 & 2 & \textcolor{gray}{0} \\
 & $z_{2,4}$ & 1.0894 & 0.6534 & \textcolor{gray}{0} & 1.1868 & \textcolor{gray}{0} & 2.3 \\
 & $z_{3,1}$ & $-$0.27 & $-$0.6057 & \textcolor{gray}{0} & $-$0.2474 & \textcolor{gray}{0} & \textcolor{gray}{0} \\
 & $z_{3,2}$ & 0.7716 & 1.1393 & \textcolor{gray}{0} & 0.8165 & 6 & \textcolor{gray}{0} \\
 & $z_{3,3}$ & $-$0.2 & $-$0.0568 & \textcolor{gray}{0} & $-$0.251 & 6 & \textcolor{gray}{0} \\
 & $z_{3,4}$ & 0.0794 & 0.515 & \textcolor{gray}{0} & 0.1318 & \textcolor{gray}{0} & \textcolor{gray}{0} \\
 & $z_{4,1}$ & 0.3903 & 0.3903 & \textcolor{gray}{0} & 0.3522 & \textcolor{gray}{0} & \textcolor{gray}{0} \\
 & $z_{4,2}$ & $-$0.1326 & $-$0.1326 & \textcolor{gray}{0} & $-$0.0391 & \textcolor{gray}{0} & \textcolor{gray}{0} \\
 & $z_{4,3}$ & 0.0156 & 0.0156 & \textcolor{gray}{0} & $-$0.0517 & \textcolor{gray}{0} & \textcolor{gray}{0} \\
 & $z_{4,4}$ & 0.2708 & 0.2708 & \textcolor{gray}{0} & 0.2345 & \textcolor{gray}{0} & \textcolor{gray}{0} \\
 & $\delta$ & 0.5253 & 0.6113 & \textcolor{gray}{0} & 0.6361 & \textcolor{gray}{0} & \textcolor{gray}{0} \\
 & $\delta_\mathrm{ref}$ & 0.3281 & 0.4134 & \textcolor{gray}{0} & 0.4844 & \textcolor{gray}{0} & \textcolor{gray}{0} \\
 & $u_{t-1}$ & 0.1651 & \textcolor{gray}{0} & \textcolor{gray}{0} & 0.0835 & \textcolor{gray}{0} & \textcolor{gray}{0} \\
 & $u_{t-2}$ & 0.0859 & \textcolor{gray}{0} & \textcolor{gray}{0} & $-$0.0592 & \textcolor{gray}{0} & \textcolor{gray}{0} \\
 & $u_{t-3}$ & $-$0.4309 & \textcolor{gray}{0} & \textcolor{gray}{0} & $-$0.2614 & \textcolor{gray}{0} & \textcolor{gray}{0} \\
 & $u_{t-4}$ & $-$0.1053 & \textcolor{gray}{0} & \textcolor{gray}{0} & $-$0.1103 & \textcolor{gray}{0} & \textcolor{gray}{0} \\
 & bias & 0.1875 & $-$0.1593 & \textcolor{gray}{0} & 0.1606 & 1.55 & $-$0.5 \\
$g_{7}$ & $z_{1,1}$ & 0.6233 & 0.6233 & \textcolor{gray}{0} & 0.4235 & \textcolor{gray}{0} & \textcolor{gray}{0} \\
 & $z_{1,2}$ & $-$0.7939 & $-$0.7939 & \textcolor{gray}{0} & $-$0.5772 & \textcolor{gray}{0} & 1.29 \\
 & $z_{1,3}$ & $-$0.1674 & $-$0.1674 & \textcolor{gray}{0} & $-$0.1931 & \textcolor{gray}{0} & \textcolor{gray}{0} \\
 & $z_{1,4}$ & $-$0.5936 & $-$0.5936 & \textcolor{gray}{0} & $-$0.3601 & \textcolor{gray}{0} & 0.85 \\
 & $z_{2,1}$ & 0.5942 & 0.3565 & \textcolor{gray}{0} & 0.548 & \textcolor{gray}{0} & \textcolor{gray}{0} \\
 & $z_{2,2}$ & $-$0.0229 & $-$0.0137 & \textcolor{gray}{0} & 0.0091 & $-$3 & 1.29 \\
 & $z_{2,3}$ & 0.3231 & 0.1939 & \textcolor{gray}{0} & 0.1204 & 2 & 0.82 \\
 & $z_{2,4}$ & $-$0.6107 & $-$0.3664 & \textcolor{gray}{0} & $-$0.7238 & \textcolor{gray}{0} & \textcolor{gray}{0} \\
 & $z_{3,1}$ & 0.173 & 0.4107 & \textcolor{gray}{0} & 0.3762 & \textcolor{gray}{0} & \textcolor{gray}{0} \\
 & $z_{3,2}$ & $-$0.1115 & $-$0.1207 & \textcolor{gray}{0} & $-$0.129 & 3 & \textcolor{gray}{0} \\
 & $z_{3,3}$ & 0.3807 & 0.5099 & \textcolor{gray}{0} & 0.2591 & 6 & \textcolor{gray}{0} \\
 & $z_{3,4}$ & $-$0.1691 & $-$0.4134 & \textcolor{gray}{0} & $-$0.3295 & \textcolor{gray}{0} & \textcolor{gray}{0} \\
 & $z_{4,1}$ & $-$0.0818 & $-$0.0818 & \textcolor{gray}{0} & $-$0.017 & \textcolor{gray}{0} & \textcolor{gray}{0} \\
 & $z_{4,2}$ & 0.1312 & 0.1312 & \textcolor{gray}{0} & 0.0978 & \textcolor{gray}{0} & \textcolor{gray}{0} \\
 & $z_{4,3}$ & $-$0.0286 & $-$0.0286 & \textcolor{gray}{0} & $-$0.1634 & \textcolor{gray}{0} & \textcolor{gray}{0} \\
 & $z_{4,4}$ & 0.2145 & 0.2145 & \textcolor{gray}{0} & 0.0823 & \textcolor{gray}{0} & \textcolor{gray}{0} \\
 & $\delta$ & $-$0.267 & $-$0.7569 & \textcolor{gray}{0} & $-$0.3499 & \textcolor{gray}{0} & $-$0.042 \\
 & $\delta_\mathrm{ref}$ & $-$0.993 & $-$1.2512 & \textcolor{gray}{0} & $-$1.141 & \textcolor{gray}{0} & $-$0.258 \\
 & $u_{t-1}$ & 0.2171 & \textcolor{gray}{0} & \textcolor{gray}{0} & 0.2373 & \textcolor{gray}{0} & \textcolor{gray}{0} \\
 & $u_{t-2}$ & 0.0283 & \textcolor{gray}{0} & \textcolor{gray}{0} & 0.1429 & \textcolor{gray}{0} & \textcolor{gray}{0} \\
 & $u_{t-3}$ & $-$0.3293 & \textcolor{gray}{0} & \textcolor{gray}{0} & $-$0.3685 & \textcolor{gray}{0} & \textcolor{gray}{0} \\
 & $u_{t-4}$ & 0.0714 & \textcolor{gray}{0} & \textcolor{gray}{0} & 0.0406 & \textcolor{gray}{0} & \textcolor{gray}{0} \\
 & bias & 0.3994 & 0.5872 & \textcolor{gray}{0} & 0.2477 & 1.55 & \textcolor{gray}{0} \\
$g_{8}$ & $z_{1,1}$ & 0.986 & 0.986 & \textcolor{gray}{0} & 0.816 & \textcolor{gray}{0} & \textcolor{gray}{0} \\
 & $z_{1,2}$ & 0.3924 & 0.3924 & \textcolor{gray}{0} & 0.5756 & \textcolor{gray}{0} & 1.29 \\
 & $z_{1,3}$ & 0.5856 & 0.5856 & \textcolor{gray}{0} & 0.5829 & \textcolor{gray}{0} & \textcolor{gray}{0} \\
 & $z_{1,4}$ & $-$0.7003 & $-$0.7003 & \textcolor{gray}{0} & $-$0.4688 & \textcolor{gray}{0} & 0.85 \\
 & $z_{2,1}$ & 0.2071 & 0.1243 & \textcolor{gray}{0} & 0.118 & \textcolor{gray}{0} & \textcolor{gray}{0} \\
 & $z_{2,2}$ & $-$1.006 & $-$0.6036 & \textcolor{gray}{0} & $-$0.8329 & \textcolor{gray}{0} & 1.29 \\
 & $z_{2,3}$ & $-$0.748 & $-$0.4488 & \textcolor{gray}{0} & $-$0.8885 & \textcolor{gray}{0} & 0.82 \\
 & $z_{2,4}$ & $-$0.7534 & $-$0.452 & \textcolor{gray}{0} & $-$0.8255 & \textcolor{gray}{0} & \textcolor{gray}{0} \\
 & $z_{3,1}$ & $-$0.036 & 0.0469 & \textcolor{gray}{0} & 0.1351 & \textcolor{gray}{0} & \textcolor{gray}{0} \\
 & $z_{3,2}$ & $-$0.1213 & $-$0.5237 & \textcolor{gray}{0} & $-$0.0818 & \textcolor{gray}{0} & \textcolor{gray}{0} \\
 & $z_{3,3}$ & 0.474 & 0.1748 & \textcolor{gray}{0} & 0.4018 & \textcolor{gray}{0} & \textcolor{gray}{0} \\
 & $z_{3,4}$ & $-$0.0325 & $-$0.3338 & \textcolor{gray}{0} & $-$0.1194 & \textcolor{gray}{0} & \textcolor{gray}{0} \\
 & $z_{4,1}$ & 0.2645 & 0.2645 & \textcolor{gray}{0} & 0.3439 & \textcolor{gray}{0} & \textcolor{gray}{0} \\
 & $z_{4,2}$ & $-$0.2417 & $-$0.2417 & \textcolor{gray}{0} & $-$0.2247 & \textcolor{gray}{0} & \textcolor{gray}{0} \\
 & $z_{4,3}$ & 0.1875 & 0.1875 & \textcolor{gray}{0} & 0.1118 & \textcolor{gray}{0} & \textcolor{gray}{0} \\
 & $z_{4,4}$ & $-$0.208 & $-$0.208 & \textcolor{gray}{0} & $-$0.221 & \textcolor{gray}{0} & \textcolor{gray}{0} \\
 & $\delta$ & $-$0.3593 & $-$0.5038 & \textcolor{gray}{0} & $-$0.4907 & \textcolor{gray}{0} & $-$0.042 \\
 & $\delta_\mathrm{ref}$ & $-$0.3831 & $-$0.4827 & \textcolor{gray}{0} & $-$0.5756 & \textcolor{gray}{0} & $-$0.258 \\
 & $u_{t-1}$ & $-$0.0034 & \textcolor{gray}{0} & \textcolor{gray}{0} & $-$0.0455 & \textcolor{gray}{0} & \textcolor{gray}{0} \\
 & $u_{t-2}$ & $-$0.0726 & \textcolor{gray}{0} & \textcolor{gray}{0} & $-$0.042 & \textcolor{gray}{0} & \textcolor{gray}{0} \\
 & $u_{t-3}$ & $-$0.0741 & \textcolor{gray}{0} & \textcolor{gray}{0} & $-$0.1727 & \textcolor{gray}{0} & \textcolor{gray}{0} \\
 & $u_{t-4}$ & $-$0.1331 & \textcolor{gray}{0} & \textcolor{gray}{0} & $-$0.1787 & \textcolor{gray}{0} & \textcolor{gray}{0} \\
 & bias & 0.1761 & 0.3567 & \textcolor{gray}{0} & 0.0841 & \textcolor{gray}{0} & \textcolor{gray}{0} \\
\midrule
$f$ & $g_{1}$ & 0.3997 & 0.8793 & 5 & 0.499 & 10 & 7.7 \\
 & $g_{2}$ & $-$1.2618 & $-$2.7764 & \textcolor{gray}{0} & $-$1.6464 & 10 & $-$7.7 \\
 & $g_{3}$ & 0.6001 & 1.3202 & \textcolor{gray}{0} & 0.8032 & 10 & 9.7 \\
 & $g_{4}$ & 0.7967 & 1.7527 & \textcolor{gray}{0} & 1.2653 & 5 & $-$9.7 \\
 & $g_{5}$ & 0.7157 & 1.5745 & \textcolor{gray}{0} & 0.9533 & $-$10 & 7 \\
 & $g_{6}$ & $-$0.7692 & $-$1.6922 & \textcolor{gray}{0} & $-$0.8165 & 5 & 7 \\
 & $g_{7}$ & 0.7443 & 1.6375 & \textcolor{gray}{0} & 0.8709 & $-$10 & 10 \\
 & $g_{8}$ & 0.6269 & 1.3792 & \textcolor{gray}{0} & 0.8855 & \textcolor{gray}{0} & 10 \\
 & bias & \textcolor{gray}{0} & \textcolor{gray}{0} & \textcolor{gray}{0} & 0.0012 & \textcolor{gray}{0} & \textcolor{gray}{0} \\
\bottomrule
\end{tabular}
\end{table*}

\clearpage

\clearpage
\bibliographystyle{cas-model2-names}
\bibliography{references}

\end{document}